%% file: main.tex
\PassOptionsToPackage{table,dvipsnames}{xcolor}
\documentclass{article}
\usepackage[OT1]{fontenc}
\usepackage{iclr2027_conference,times}

\usepackage{eccvabbrv}

\usepackage{graphicx}
\usepackage{booktabs}
\usepackage{caption}
\usepackage{microtype}
\input{supp-preamble}

\DeclareCaptionJustification{balanced}{\leftskip=0pt\rightskip=0pt\parfillskip=0pt plus .25\linewidth\relax}
\usepackage{iftex}
\ifPDFTeX
  \usepackage[accsupp]{axessibility}  %
\else
  \usepackage{amsmath,amssymb}
\fi

\usepackage{hyperref}
\usepackage{url}
\usepackage{xurl} %

\usepackage{multirow}
\usepackage[table,dvipsnames]{xcolor} %
\usepackage{siunitx}

\newcommand{\rev}[1]{\textcolor{blue}{#1}}
\usepackage{arxiv-preprint}

\begin{document}

\title{{\fontsize{15.5}{20}\selectfont\reasonuniname: Reasoning-Augmented Reconstruction\\Advances Unified Multimodal Models}}
\hypersetup{pdftitle={ReasonUni: Reasoning-Augmented Reconstruction Advances Unified Multimodal Models}}

\input{authors}

\maketitle
\vspace{-12pt}
\setlength{\parfillskip}{0pt plus .25\linewidth}

\begin{abstract}
\setlength{\parfillskip}{0pt plus .25\linewidth}

\textcolor{ForestGreen}{Image-to-text (I2T) understanding and text-to-image (T2I) generation are dual tasks in unified multimodal models (UMMs). Reconstructive post-training couples these tasks, but existing reconstruction-based methods remain perception-centric and provide limited supervision for reasoning about structure, intent, and transformation. We propose \texttt{\textbf{ReasonUni}}, a reasoning-augmented reconstructive post-training framework that extends this coupling beyond perception.}
\textcolor{ForestGreen}{Given an original image and an editing instruction, the UMM first reasons about \textit{what} and \textit{how} to edit and describes the target state, then generates the corresponding target image conditioned on this understanding output.}
\textcolor{ForestGreen}{To jointly optimize both modules, we introduce \texttt{\textbf{TreeGRPO}}, a tree-structured reinforcement learning method that samples multiple reasoning paths and candidate images per path, then aggregates visual rewards to update both modules under a unified objective. Experiments across multiple benchmarks demonstrate consistent gains in both understanding and generation, showing that reasoning-augmented reconstruction yields genuine mutual gains for UMMs.}

\end{abstract}
\vspace{-6pt}

\noindent\begin{minipage}{\linewidth}
    \centering
    \includegraphics[width=\linewidth]{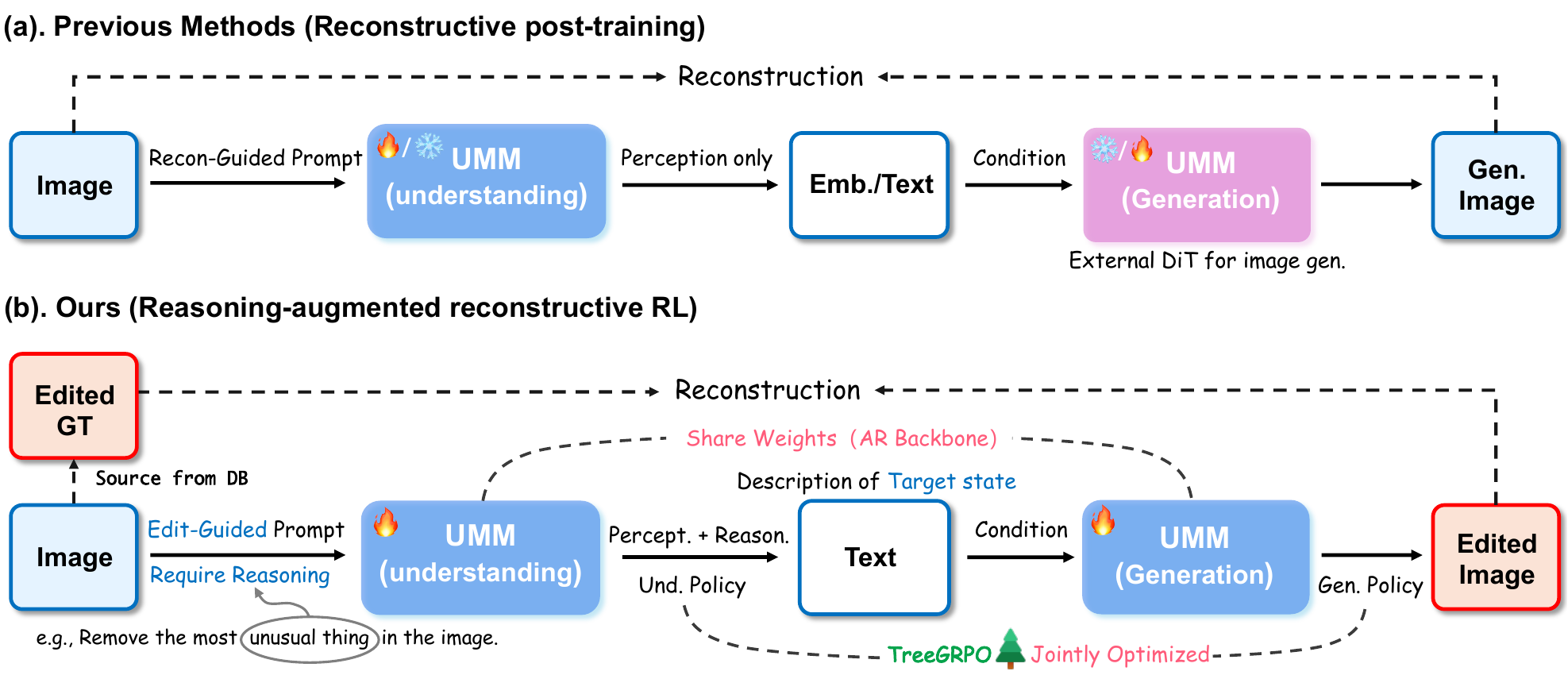}
    \captionsetup{skip=4pt}
    \captionof{figure}{\textcolor{blue}{\textbf{From perception-only to reasoning-augmented reconstruction.} (a) Previous methods reconstruct the observed image. (b) ReasonUni infers an edited target state and uses TreeGRPO to jointly optimize understanding and generation through reconstruction feedback.}}
    \label{fig:motivation}
\end{minipage}\par

\section{Introduction}
\label{sec:intro}

Unified Multimodal Models (UMMs)~\cite{wang2024emu3, chen2025janus, wu2025janus, xie2024show, pan2025metaquery, tong2025metamorph, zhou2024transfusion, yan2025gpt} have been proposed to both understand and generate across modalities--generating visual content and text within a single unified architecture.
However, naively coupling understanding and generation often yields suboptimal results: prior UMM studies~\cite{pan2025metaquery, chen2025blip3} report that optimizing diffusion-based generative objectives can impair understanding performance and learned representations (and vice versa).

\begin{figure}[!t]
    \centering
    \includegraphics[width=\linewidth]{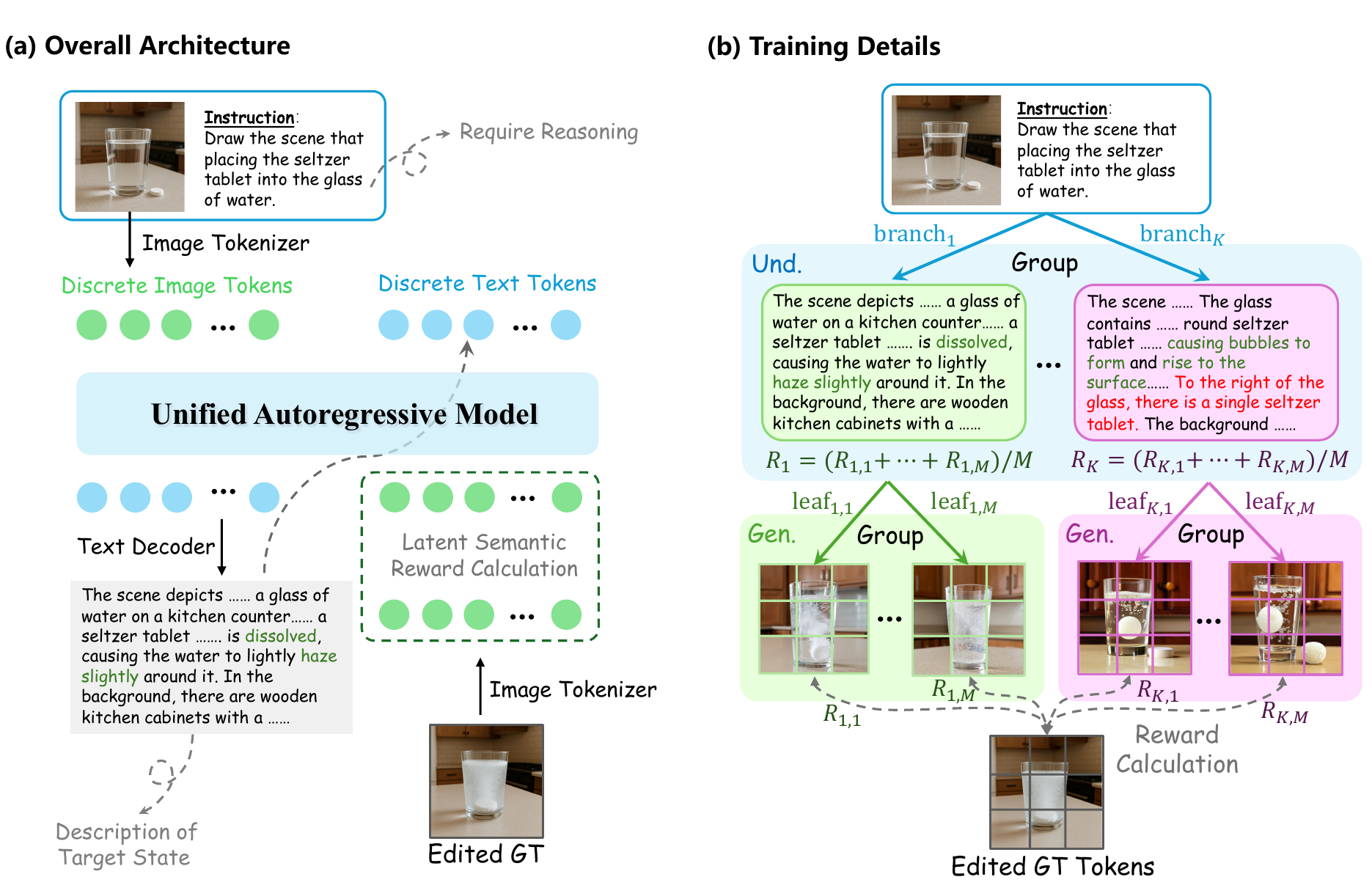}
    \caption{Overview of the ReasonUni architecture and method (left). Given an input image and reasoning-required editing instruction, the UMM infers and describes the target state, then generates candidate edited images conditioned on these descriptions. TreeGRPO formulates this process as a hierarchical branch--leaf structure: reasoning paths form branches, generated images form leaves, and visual rewards are aggregated and propagated to optimize understanding and generation jointly (right).}
    \label{fig:pipeline}
\end{figure}

\textcolor{ForestGreen}{To address these conflicts, Janus~\cite{wu2025janus} separates visual encoders for understanding and generation, while TokenFlow~\cite{qu2025tokenflow} aligns semantic and pixel-level codebooks through shared indices.}
\textcolor{ForestGreen}{However, architectural unification alone does not establish mutual gains between understanding and generation, motivating training strategies that explicitly couple the two tasks.}

Rather than redesigning the architecture, recent works have explored an alternative direction: dedicated \textit{post-training} strategies to advance UMMs and explore potential mutual benefits~\cite{chen2025blip3o,su2026generation,yan2025uae,xie2025reca}.
One promising instantiation is to frame UMMs as auto-encoders and post-train both modules with a reconstruction objective~\cite{yan2025uae,xie2025reca}.
To illustrate, RecA~\cite{xie2025reca} uses embeddings from the visual understanding module as dense captions to condition text-to-image generation, yielding consistent gains across benchmarks. UAE~\cite{yan2025uae} models the UMM as an "Image--Text--Image" auto-encoder: the encoder captions visual details, and the decoder reconstructs the image from the caption. By optimizing reconstruction, UAE improves generation quality and also strengthens perceptual understanding of fine-grained details.

However, the gains in understanding are still limited at the "perception" level~\cite{yan2025uae,wang2024reconstructive}, as the prior reconstruction frameworks emphasize only recognizing what is present, without requiring the model to reason anything about structure, intent, or causal relations in the scene. \textcolor{ForestGreen}{As a result, their ability to generalize to complex, open-ended tasks that require higher-order visual reasoning remains limited.}

To bridge this gap, we propose ReasonUni, a reasoning-augmented reconstructive post-training method that extends supervision beyond perception to visual reasoning. Given an input image and an editing instruction, the UMM first reasons about the required transformation and produces a detailed target description of the desired edits. The generation module then reconstructs the target image conditioned on the description.

\textcolor{ForestGreen}{To jointly optimize both modules, we propose TreeGRPO, a hierarchical tree-based framework that enables cleaner credit assignment between understanding and generation. Each sampled reasoning path forms a branch with multiple generated images as leaves. Aggregating their rewards provides a more reliable branch-level signal for understanding, while within-branch normalization guides generation under a fixed target-state description. Applying Group Relative Policy Optimization (GRPO)~\cite{shao2024deepseekmath} at these two levels jointly optimizes both modules under a unified objective, rather than optimizing them in separate stages~\cite{yan2025uae}.}

To make this joint optimization practical, we adopt a more unified version of the UMM series (i.e., X-Omni~\cite{geng2025x} style) with a pure AR backbone that generates both image and text tokens, so image and text generation are optimized under one AR objective. In contrast, note that prior work like UAE mainly uses a UniWorld-style~\cite{lin2025uniworld} split design (AR for text with an external diffusion image generator), and only the AR component is directly optimized.

\textcolor{ForestGreen}{Experiments demonstrate consistent gains in both understanding and generation across multiple benchmarks. Compared with UAE, ReasonUni delivers larger gains in both capabilities, outperforming it by up to 5.98 points on DPGBench, 13.00 on WISE, and 16.78 on GenEval\raisebox{.4ex}{\tiny ++}. On MMBench, ReasonUni improves both perception and reasoning, with logical reasoning gaining 6.77 points over X-Omni versus 3.38 for UAE. These understanding gains extend across all 12 evaluated benchmarks, supporting the broad effectiveness of our approach.}

\section{Method}
\subsection{Preliminaries: Unified Discrete Autoregressive Modeling}
We build on a unified discrete autoregressive (AR) multimodal model (UMM), instantiated with X-Omni~\cite{geng2025x}, with parameters $\theta$. The key design is a shared \textit{discrete} token space for both image and text. Specifically, an image is quantized into visual tokens by a visual tokenizer, and text is represented as language tokens. Importantly, visual inputs (for understanding) and visual outputs (for generation) use the same semantic codebook, so both directions are aligned at the representation level.

Due to this unified and simplified architecture design, the UMM can be trained with standard next-token AR likelihood. A separate decoder can map visual tokens back to RGB, but semantic layout and content are modeled in the latent token space by AR. Under this formulation, one backbone \textit{naturally} supports two conditional policies: understanding $p_\theta(T\mid V)$ and generation $p_\theta(V\mid T)$. This dual-policy view is the foundation of our method: by chaining the two policies in sequence, we can construct training signals that flow from image-level feedback back to the reasoning process, enabling true end-to-end joint optimization.

\subsection{ReasonUni: Reasoning-Augmented Reconstructive Framework for UMMs}
Conventional reconstructive post-training~\cite{yan2025uae,xie2025reca} improves semantic completeness for UMMs, but often emphasizes perception more than visual reasoning. In practice, the model may learn to preserve or copy content, while receiving weak supervision on \textit{what to change}, \textit{how to change it}, and \textit{what to keep unchanged}, \textcolor{ForestGreen}{leaving the model's higher-order visual reasoning abilities largely unsupervised.}

To address this, we propose \textbf{ReasonUni}, which turns reconstruction into a reasoning-driven editing task: the model must interpret an editing instruction, reason about its implications, and produce a coherent target description before generating the edited image. For each training tuple $(x_{src}, c, x_{gt})$, where $x_{src}$ is the source image, $c$ is an edit instruction, and $x_{gt}$ is the target image, we apply two coupled steps.

{\color{ForestGreen}
\noindent\textbf{\textit{Step 1: Understanding.}}
The understanding module first predicts a textual description $y$ of the desired target state: $y \sim p_\theta(\cdot \mid x_{src}, c).$ To produce $y$, the module must infer the edit required by the instruction, including implicit targets and transformations, while perceiving the objects, attributes, and spatial relations in the source image that should remain unchanged. The resulting description therefore integrates the module's reasoning about the intended edit with its perception of the source scene. The model directly outputs a description of the edited scene rather than an explicit chain-of-thought trace; the prompting template is provided in Appendix~\ref{sec: prompt_templates}.
}

\noindent\textbf{\textit{Step 2: Generation.}}
The generation module then synthesizes an edited image conditioned on \textcolor{ForestGreen}{the target-state description}: $\hat{x} \sim p_\theta(\cdot \mid y).$ Because generation receives only $y$ (not the source image directly), \textcolor{ForestGreen}{faithful reconstruction requires the description to capture both the intended edit and the scene content to be preserved}.

This coupling is central: image-level feedback supervises not only generation quality but also the quality of the upstream reasoning signal, so both components receive learning signals accordingly.

\subsection{TreeGRPO: Tree-based Joint Group Relative Policy Optimization}
To jointly optimize understanding and generation, we introduce \textbf{TreeGRPO}. Standard GRPO~\cite{shao2024deepseekmath} samples multiple outputs for a single input and uses relative comparisons to compute advantages. \textcolor{ForestGreen}{However, our two-stage pipeline introduces variation in both the target-state description and the image generated from it.} A flat sampling strategy would conflate these two sources, making credit assignment difficult.

\begin{figure}[t]
    \centering
    \includegraphics[width=\linewidth]{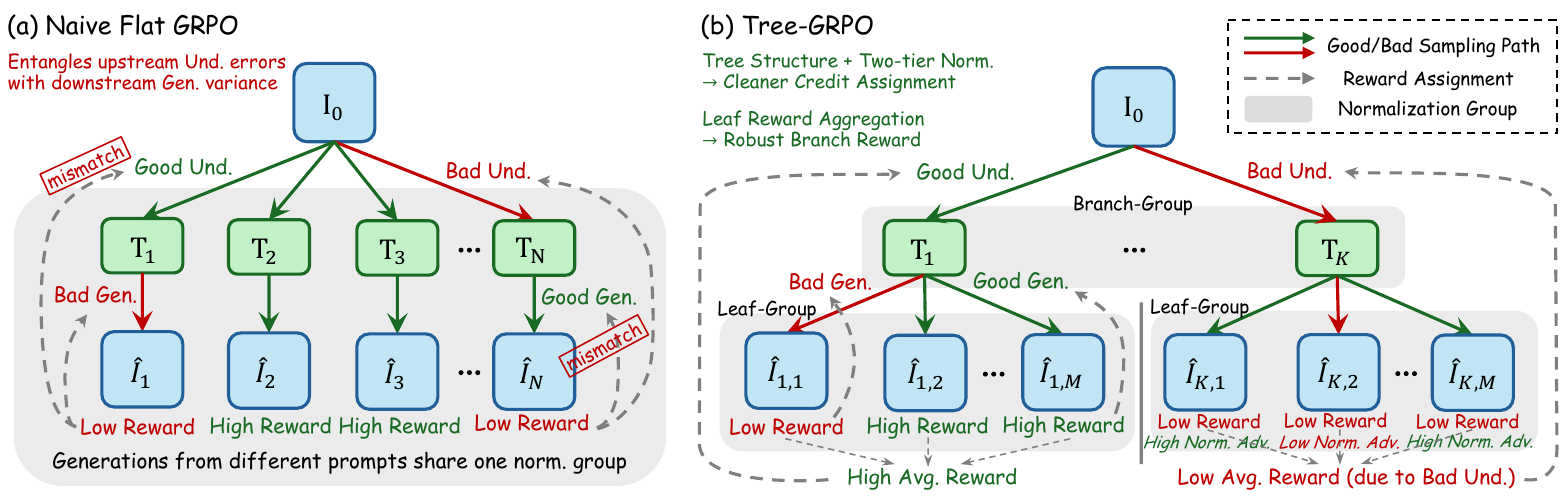}
    \caption{\textcolor{ForestGreen}{Flat vs.\ tree-structured sampling. \textbf{Left (Flat):} each reasoning path and its generated image form an independent sample. \textbf{Right (Tree):} each of $K$ reasoning paths branches into $M$ generated images, enabling group-relative comparisons across branch-level mean rewards and among image rewards within each branch. Even when a branch has a low mean reward, images scoring above that branch mean receive positive generation advantages.}}
    \label{fig:flat-vs-tree}
\end{figure}

\textcolor{ForestGreen}{As illustrated in Figure~\ref{fig:flat-vs-tree}, TreeGRPO organizes candidate outputs into a tree, with reasoning paths as branches and generated images as leaves. Comparing mean rewards across branches supervises understanding, while comparing individual image rewards within each branch guides image generation under a fixed target-state description.}

\noindent\textbf{Hierarchical sampling.}
\textcolor{ForestGreen}{Given $(x_{src}, c, x_{gt})$, we sample $K$ reasoning paths from the understanding module and $M$ edited image candidates from the generation module for each path:}
\begin{equation}
    y_k \sim p_\theta(\cdot \mid x_{src}, c), \quad
    \hat{x}_{k,j} \sim p_\theta(\cdot \mid y_k),
    \quad k=1,\dots,K,\ j=1,\dots,M.
\end{equation}
This produces a tree with $K$ branches and $K \times M$ leaves. The two-level sampling separates uncertainty at two levels: \textit{which reasoning path is best} and \textit{how well each path can be rendered}, with multiple images per path providing a robust estimate of each path's quality.

\noindent\textbf{Latent semantic reward.}
\textcolor{ForestGreen}{Unlike UAE~\cite{yan2025uae}, which computes reconstruction rewards using CLIP features of decoded images, we instead compute reconstruction rewards directly in X-Omni's semantically aligned visual token space, which effectively captures high-level semantic similarity while remaining lightweight and efficient.}

\textcolor{ForestGreen}{Let $q(\cdot)$ denote the visual tokenizer and $e(v_i)$ the codebook embedding associated with visual token $v_i$. SigLIP-VQ~\citep{geng2025x} learns a visual codebook by quantizing contextualized patch features from a SigLIP encoder. Positional embeddings and self-attention incorporate spatial context into these features before quantization, allowing the learned token representations to encode spatial cues alongside semantic content. To compute the reward, we mean-pool the codebook embeddings of the generated tokens and the tokenized target image, then measure the cosine similarity between the two representations. The pooled representations can thus retain implicit spatial cues even after this aggregation.} \rev{Operating directly in the model-native latent space also avoids decoding tokens into pixels and re-encoding them with an external vision encoder, thereby reducing information loss and representation mismatch.}
\textcolor{ForestGreen}{Let $\hat{\mathbf v}_{k,j}$ denote the generated visual token sequence corresponding to $\hat{x}_{k,j}$.}
\begin{equation}
\color{ForestGreen}
\begin{aligned}
h(\mathbf v) &= \frac{1}{L}\sum_{i=1}^{L} e(v_i), \qquad \mathbf v=(v_1,\ldots,v_L),\\
R_{k,j} &= \operatorname{CosSim}\!\left(h(\hat{\mathbf v}_{k,j}), h(q(x_{gt}))\right).
\end{aligned}
\end{equation}

\noindent\textbf{Group-relative advantages.}
The key insight is that the two modules should be evaluated at different granularities: the understanding module is rewarded based on the average quality of all images from its reasoning path, while the generation module is compared against siblings under the same path. We aggregate leaf rewards into branch scores, then standardize at two levels:
\begin{equation}
\begin{aligned}
S_k &= \frac{1}{M}\sum_{j=1}^{M} R_{k,j}, \qquad
A_k^{und} &= \frac{S_k-\mu^{und}}{\sigma^{und}+\epsilon}, \qquad
A_{k,j}^{gen} = \frac{R_{k,j}-\mu_k^{gen}}{\sigma_k^{gen}+\epsilon}.
\end{aligned}
\end{equation}
Here $(\mu^{und},\sigma^{und})$ are computed over branch scores $\{S_k\}_{k=1}^{K}$, while $(\mu_k^{gen},\sigma_k^{gen})$ are computed over candidate rewards $\{R_{k,j}\}_{j=1}^{M}$ within branch $k$. This two-level normalization ensures each module is compared only against alternatives at its own level of the hierarchy.

\noindent\textbf{Unified tree objective.}
\textcolor{ForestGreen}{Both modules are jointly optimized under a single objective, with each branch contributing an understanding term and a generation term:}
\begin{equation}
    \mathcal{J}(\theta) = \frac{1}{K} \sum_{k=1}^{K} \left[ \mathcal{J}_{und}(y_k;A_k^{und}) + \lambda_{gen}\frac{1}{M}\sum_{j=1}^{M} \mathcal{J}_{gen}(\hat{x}_{k,j}\mid y_k;A_{k,j}^{gen}) \right],
\end{equation}
where $\lambda_{gen}$ controls the relative weight of generation updates (set to $1$ in our experiments).

\textcolor{ForestGreen}{Both objectives use token-level GRPO with KL regularization to reference policy $\pi_{ref}$:}
\begin{equation}
\begin{aligned}
\mathcal{J}_{token}(\theta) = \mathbb{E}_{t}\Big[&\min\big(r_tA_t,\operatorname{clip}(r_t,1-\epsilon_c,1+\epsilon_c)A_t\big) \\
&- \beta D_{KL}\big(\pi_\theta(\cdot\mid o_{<t})\,\|\,\pi_{ref}(\cdot\mid o_{<t})\big)\Big],
\end{aligned}
\end{equation}
where $r_t = \frac{\pi_\theta(o_t\mid o_{<t})}{\pi_{\theta_{old}}(o_t\mid o_{<t})}$. Here $o_t$ is a token from the shared vocabulary, and $A_t$ is assigned from $A_k^{und}$ for \textcolor{ForestGreen}{text tokens in $y_k$} or from $A_{k,j}^{gen}$ for \textcolor{ForestGreen}{visual tokens in $\hat{\mathbf v}_{k,j}$}. The clipping mechanism and KL penalty jointly maintain training stability.

\section{Experiments}

\textcolor{ForestGreen}{We evaluate whether reasoning-augmented reconstruction improves understanding and generation jointly, rather than trading one capability for the other. Our experiments address three questions:}
\rev{(Q1) Can ReasonUni achieve simultaneous improvements in both understanding and generation across different initialization settings?}
(Q2) Are the gains robust across diverse benchmark families?
\rev{(Q3) Which aspects of understanding and generation contribute to the gains?}

\textcolor{ForestGreen}{We train on a mixture of existing editing datasets and synthetic reasoning-driven editing data. Visual understanding is evaluated using VLMEvalKit~\citep{duan2024vlmevalkit} with a decoding temperature of 0. Details of data synthesis, training configurations,\linebreak and evaluation protocols are provided in Appendices~\ref{app:data-synthesis}--\ref{app:implementation}.}

\subsection{Main Quantitative Results}
\rev{To answer Q1, Table~\ref{tab:main_results} compares ReasonUni with X-Omni and UAE~\cite{yan2025uae} from the X-Omni-SFT and X-Omni-RL checkpoints, which we refer to as the SFT and RL settings, respectively. \textcolor{ForestGreen}{UAE is reimplemented on X-Omni for both settings, with implementation details and matched-data comparison settings provided in Appendix~\ref{app:baseline-details}.}}
\textcolor{ForestGreen}{ReasonUni achieves simultaneous gains in understanding and generation across both initialization settings, outperforming X-Omni and UAE on all seven reported metrics. In the SFT setting, generation improves over X-Omni by 13.34 points on DPGBench, 13.00 on WISE, and 15.71 on GenEval\raisebox{.4ex}{\tiny ++}, alongside gains of 0.76--4.00 points on the four understanding benchmarks. The advantage also holds across all seven metrics in the matched-data control (UAE$^\dagger$), indicating that the gains are not simply due to training data. The same joint improvement holds in the RL setting, including gains over X-Omni of 1.65 on MMBench, 2.00 on MMVP, 1.75 on CV-Bench, and 6.79 on GenEval\raisebox{.4ex}{\tiny ++}. Together, these results establish consistent mutual gains across initialization settings, rather than improving one capability at the expense of the other.}

\textcolor{ForestGreen}{Unless otherwise noted, subsequent experiments use the RL setting. To answer Q2, Table~\ref{tab:muti-bench-comparison} extends the understanding evaluation to 12 metrics across six benchmark families. ReasonUni improves over X-Omni on every metric, with the largest gains on MMMU (+2.67), MMVP (+2.00), and CV-Bench (+1.75). These consistent gains across diverse benchmarks provide broader evidence that ReasonUni improves visual understanding.}

\begin{table}[tb] %
  \centering
  {\fontsize{8.5}{10.2}\selectfont
    \setlength{\tabcolsep}{7.03pt}
    \renewcommand{\arraystretch}{1.15} %
    \begin{tabular}{l cccc ccc}
    \toprule
    \multirow{2}{*}{\textbf{Method}} & \multicolumn{4}{c}{\textbf{Understanding}} & \multicolumn{3}{c}{\textbf{Generation}} \\
    \cmidrule(lr){2-5} \cmidrule(lr){6-8}
    & MMBench & MMVP & CV-Bench  & MathVerse & DPGBench & WISE & GenEval\raisebox{.4ex}{\tiny ++} \\
    \midrule

    X-Omni-SFT      & 74.40 & 48.67 & 67.98  & 17.01 & 65.25 & 24.00 & 7.50 \\
    + UAE           & 74.40 & 50.67 & 67.06  & 17.51 & 72.61 & 24.00 & 6.43 \\
    \textcolor{gray!65}{+ UAE$^\dagger$} & \textcolor{gray!65}{75.26} & \textcolor{gray!65}{51.33} & \textcolor{gray!65}{66.87} & \textcolor{gray!65}{17.64} & \textcolor{gray!65}{65.49} & \textcolor{gray!65}{26.00} & \textcolor{gray!65}{6.43} \\
    \rowcolor{gray!15}\textbf{+ Ours} & \textbf{75.60} & \textbf{52.67} & \textbf{68.85}  & \textbf{17.77} & \textbf{78.59} & \textbf{37.00} & \textbf{23.21} \\

    \textit{Gain ($\uparrow$)} & \textcolor{GainGreen}{+1.20} & \textcolor{GainGreen}{+4.00} & \textcolor{GainGreen}{+0.87}  & \textcolor{GainGreen}{+0.76} & \textcolor{GainGreen}{+13.34} & \textcolor{GainGreen}{+13.00} & \textcolor{GainGreen}{+15.71} \\

    \midrule

    X-Omni-RL       & 74.31 & 50.67 & 68.26 & 17.39 & 82.29 & 45.00 & 40.00 \\
    + UAE           & 75.43 & 50.00 & 68.05  & 17.51 & 83.32 & 43.00 & 38.57 \\

    \rowcolor{gray!15}\textbf{+ Ours} & \textbf{75.96} & \textbf{52.67} & \textbf{70.01} & \textbf{18.40} & \textbf{83.58} & \textbf{48.00} & \textbf{46.79} \\

    \textit{Gain ($\uparrow$)} & \textcolor{GainGreen}{+1.65} & \textcolor{GainGreen}{+2.00} & \textcolor{GainGreen}{+1.75}  & \textcolor{GainGreen}{+1.01} & \textcolor{GainGreen}{+1.29} & \textcolor{GainGreen}{+3.00} & \textcolor{GainGreen}{+6.79} \\
    \bottomrule
    \end{tabular}%
  }
  \vspace{1mm}
  \caption{\textbf{Quantitative comparison on multimodal understanding and generation.} \rev{We benchmark our method against X-Omni and UAE~\cite{yan2025uae} under the SFT and RL initialization settings. The gray $\dagger$ row uses the same training data as ReasonUni, while the unmarked UAE rows use UAE's original training data.} \textbf{Bold} denotes the best results. Gain indicates the absolute improvement over the corresponding X-Omni baseline.}
  \label{tab:main_results}
\end{table}

\begin{table}[tb]
  \centering
  {\fontsize{7.5}{9}\selectfont
    \setlength{\tabcolsep}{1.5pt}
    \renewcommand{\arraystretch}{1.3}
    \begin{tabular}{l cccc ccc cc c c c}
    \toprule
    \multirow{2}{*}{\textbf{Method}} & \multicolumn{4}{c}{\textbf{General}} & \multicolumn{3}{c}{\textbf{Vision-Centric}} & \multicolumn{2}{c}{\textbf{Hallucination}} & \textbf{OCR} & \textbf{Knowledge} & \textbf{Math} \\
    \cmidrule(lr){2-5} \cmidrule(lr){6-8} \cmidrule(lr){9-10} \cmidrule(lr){11-11} \cmidrule(lr){12-12} \cmidrule(lr){13-13}
    & MMB & SEED & MMStar & MMT & VStar & MMVP & CVB. & POPE & Hallu. & OCRB. & MMMU & MathV. \\
    \midrule
    Show-o~\cite{xie2024show}    & 47.9 & --    & --    & --    & --    & 50.0  & --    & 73.8  & --    & --    & 25.1  & -- \\
    Emu3~\cite{wang2024emu3}      & 58.5  & 68.2  & --    & --    & 51.8  & 36.6  & --    & 85.2  & --    & 68.7  & 31.6  & -- \\
    VILA-U~\cite{wu2024vila}    & 66.6  & 57.1  & --    & --    & 38.7  & 22.0  & --    & 85.8  & --    & --    & 32.2  & -- \\
    Janus Pro~\cite{chen2025janus} & 79.2  & 72.1  & --    & --    & --    & --    & --    & 87.4  & --    & 59.5  & 41.0  & -- \\
    MetaMorph~\cite{tong2025metamorph} & 75.2  & 71.8  & --    & --    & 44.0  & 48.3  & --    & --    & --    & --    & 41.8  & -- \\
    Mogao~\cite{liao2025mogao}     & 75.0  & 74.6  & --    & --    & --    & --    & --    & 88.9  & --    & --    & 44.2  & -- \\
    Bagel~\cite{bagel}     & 79.2  & --    & --    & --    & --    & 54.7  & --    & --    & --    & --    & 43.2  & -- \\
    \midrule
    X-Omni-RL    & 74.31 & 74.04 & 52.60 & 57.08 & 48.68 & 50.67 & 68.26 & 88.98 & 60.46 & 70.9  & 37.33    & 17.39 \\
    \rowcolor{gray!15}+Ours     & 75.96 & 74.14 & 53.27 & 57.40 & 50.26 & 52.67 & 70.01 & 89.13 & 61.30 & 71.6  & 40.00    & 18.40 \\
    \midrule
    Gain ($\uparrow$) & \textcolor{GainGreen}{+1.65} & \textcolor{GainGreen}{+0.10} & \textcolor{GainGreen}{+0.67} & \textcolor{GainGreen}{+0.32} & \textcolor{GainGreen}{+1.58} & \textcolor{GainGreen}{+2.00} & \textcolor{GainGreen}{+1.75} & \textcolor{GainGreen}{+0.15} & \textcolor{GainGreen}{+0.84} & \textcolor{GainGreen}{+0.70} & \textcolor{GainGreen}{+2.67} & \textcolor{GainGreen}{+1.01} \\
    \bottomrule
    \end{tabular}
  }
\vspace{1mm}
\captionsetup{font=small, labelfont=bf}
\caption{\textbf{Comparison with state-of-the-art unified models.} We conduct a comprehensive evaluation across six domains, including \textbf{General} (MMBench~\cite{liu2023mmbench}, SEED-Bench~\cite{li2023seed}, MMStar~\cite{chen2024mmstar}, MMT~\cite{mmt}), \textbf{Vision-Centric} (VStar~\cite{wu2024vstar}, MMVP~\cite{tong2024mmvp}, CV-Bench~\cite{tong2024cvbench}), \textbf{Hallucination} (POPE~\cite{li2023pope}, HallusionBench~\cite{guan2024hallusionbench}), \textbf{OCR} (OCRBench~\cite{liu2024ocrbench}), \textbf{Knowledge} (MMMU)~\cite{yue2024mmmu}, and \textbf{Math} (MathVerse~\cite{zhang2024mathverse}). -- denotes results not reported in original papers.}
\label{tab:muti-bench-comparison}
\end{table}

\begin{table*}[tb]
  \centering
  {\fontsize{8.5}{10.2}\selectfont
  \setlength{\tabcolsep}{7.43pt}
  \renewcommand{\arraystretch}{1.15} %
  \begin{tabular}{l ccc ccc c}
    \toprule
    \multirow{2}{*}{\textbf{Model}} & \multicolumn{3}{c}{\textbf{Perception}} & \multicolumn{3}{c}{\textbf{Reasoning}} & \multirow{2}{*}{\textbf{Overall}} \\
    \cmidrule(lr){2-4} \cmidrule(lr){5-7}
    & CoarseP. & FP-Cross & FP-Single  & AttributeR. & LogicalR. & RelationR. \\
    \midrule
    X-Omni-RL & 84.12 & 66.43 & 77.13 & 76.88 & 48.31 & 73.91 & 74.31\\
    +UAE & 84.46 & 67.13 & \textbf{79.18} & 77.39 & 51.69 & 73.91 & 75.43\\
    \rowcolor{gray!15}\textbf{+Ours} & \textbf{84.46} & \textbf{67.83} & 78.50 & \textbf{78.39} & \textbf{55.08} & \textbf{74.78} & \textbf{75.96} \\
    \midrule
    Gain($\uparrow$) & \textcolor{GainGreen}{+0.34} & \textcolor{GainGreen}{+1.40} & \textcolor{GainGreen}{+1.37} & \textcolor{GainGreen}{+1.51} & \textcolor{GainGreen}{+6.77} & \textcolor{GainGreen}{+0.87} & \textcolor{GainGreen}{+1.65} \\
    \bottomrule
  \end{tabular}
  }
  \vspace{1mm}
  \caption{\textbf{Fine-grained analysis on MMBench.} We report detailed performance across \textbf{Perception} (Coarse-Perception, Fine-grained Cross-instance, Fine-grained Single-instance) and \textbf{Reasoning} (Attribute Reasoning, Logical Reasoning, Relation Reasoning) sub-tasks. Our method improves both perception and reasoning capabilities, and the gains in reasoning are more significant.}
  \label{tab:mmbench_performance}
\end{table*}

\begin{table}[htbp]
  \centering

  {\fontsize{8.5}{10.2}\selectfont
    \renewcommand{\arraystretch}{1.15} %
    \setlength{\tabcolsep}{4.21pt}

    \begin{tabular}{l ccc ccc cc}
    \toprule
    \textbf{Model} & \textbf{Color} & \textbf{Pos/Cnt} & \textbf{Col/Pos} & \textbf{Col/Cnt} & \textbf{Mul-Cnt} & \textbf{Pos/Size} & \textbf{Count} & \textbf{Overall} \\
    \midrule
    Janus-Pro~\cite{chen2025janus} & 45.0 & 7.5  & 30.0 & 12.5 & 12.5 & 35.0 & 30.0 & 24.60 \\
    T2I-R1~\cite{jiang2025t2i}       & 67.5 & 7.5  & 35.0 & 20.0 & 30.0 & 25.0 & 32.5 & 31.10 \\
    BLIP3o~\cite{chen2025blip3}      & 25.0 & 12.5 & 60.0 & 12.5 & 22.5 & 57.5 & 25.0 & 30.70 \\
    OmniGen2~\cite{wu2025omnigen2}   & 55.0 & 12.5 & 27.5 & 20.0 & 45.0 & 25.0 & 42.5 & 32.50 \\
    Bagel~\cite{bagel}               & 32.5 & 25.0 & 32.5 & 25.0 & 37.5 & 47.5 & 60.0 & 37.10 \\
    \midrule
    X-Omni-RL                        & 25.0 & 32.5 & 47.5 & 15.0 & 45.0 & 55.0 & 60.0 & 40.00 \\
    \rowcolor{gray!15}+Ours                  & 42.5 & 35.0 & 50.0 & 12.5 & 47.5 & 80.0 & 60.0 & 46.79 \\
    \bottomrule
    \end{tabular}%
  }
  \vspace{1mm}
  \caption{\textbf{Performance comparison on GenEval\raisebox{.4ex}{\tiny ++}.} This benchmark evaluates fine-grained image generation capabilities across compositional attributes (e.g., Color, Position, Count). \textcolor{ForestGreen}{Our method achieves the highest overall score among the compared models.}}
  \label{tab:geneval_subtask_comparison}
\end{table}

\subsection{Fine-grained Quantitative Analysis}
\textcolor{ForestGreen}{We next examine the understanding and generation subcategories behind these gains (Q3).}

\textbf{Understanding side.}
On MMBench, gains are strongest in reasoning-related dimensions rather than perception-only dimensions (Table~\ref{tab:mmbench_performance}).
Compared with X-Omni, ReasonUni improves all six dimensions, with the largest gain on logical reasoning (LR: 55.08 vs.
48.31, +6.77).
Compared with UAE, ReasonUni keeps perception comparable (CP tied at 84.46; FP-C +0.70) while providing consistent reasoning gains (AR +1.00, LR +3.39, RR +0.87), improving overall MMBench from 75.43 to 75.96.
This suggests the main benefit is not merely better recognition, but stronger higher-order reasoning.

\textbf{Generation side.}
\rev{On GenEval\raisebox{.4ex}{\tiny ++} (Table~\ref{tab:geneval_subtask_comparison}), ReasonUni achieves the best overall score of 46.79, improving over X-Omni-RL by +6.79 and outperforming UAE baseline. The gains are most pronounced on Pos/Size (+25.0) and Color (+17.5), with consistent improvements on Pos/Count, Color/Position, and Multi-Count (+2.5 each). These results demonstrate stronger control over attributes, spatial relations, and multi-object composition.}

\subsection{\rev{Ablation Study}}
\label{sec:ablation}

\begin{figure}[t]
  \centering
  \includegraphics[width=1\linewidth]{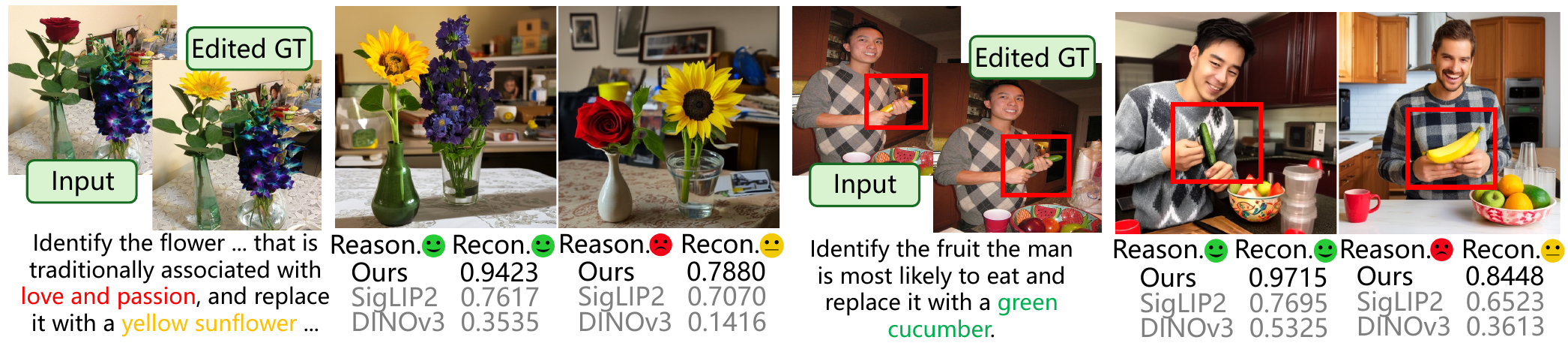}
  \caption{\textcolor{ForestGreen}{\textbf{Comparison of reconstruction rewards.} Across these examples, the latent reward produces the same preference ordering over candidate reconstructions as SigLIP2 and DINOv3, while operating directly on visual tokens without image decoding or external re-encoding.}}
  \label{fig:reward-compare}
\end{figure}

\noindent\rev{\textbf{Reward design.}
Latent rewards improve training efficiency without sacrificing performance: Table~\ref{tab:ablation_reward}(a) shows the highest Und.\,Avg (54.26) and Gen.\,Avg (47.40) among the compared rewards, together with a $\sim$60\% reduction in training time (25h vs.\ 58--68h on 8$\times$H100). Thus, the efficiency gain accompanies stronger understanding and generation, rather than a quality--cost trade-off. We compute rewards directly in SigLIP-VQ token space, bypassing diffusion decoding and external re-encoding. \textcolor{ForestGreen}{This design avoids diffusion-induced noise and external-encoder mismatch, while using SigLIP-VQ codebook embeddings that can implicitly retain spatial cues.} \textcolor{ForestGreen}{Figure~\ref{fig:reward-compare} provides qualitative evidence that the latent reward captures task-relevant differences in reconstruction quality consistently with external visual encoders.}}

\noindent\textcolor{ForestGreen}{\textbf{Sampling strategy.}
TreeGRPO makes more effective use of the image-sampling budget than flat GRPO (Table~\ref{tab:ablation_reward}(b)). With the same 16 candidate images, our default $K{=}4,M{=}4$ configuration, highlighted in gray, improves Und.\,Avg from 53.90 to 54.26 and Gen.\,Avg from 45.40 to 47.40, while reducing training time from 30h to 25h. These joint gains demonstrate the value of tree-structured optimization at a fixed image-sampling budget. Sharing each description across four images requires only four target-state descriptions instead of sixteen, reducing text-generation overhead. Scaling to $K{=}8,M{=}8$ further increases Gen.\,Avg to 49.02, showing that generation benefits from a larger tree-sampling budget.}

\begin{table}[t]
  \centering
  \fontsize{7.5}{9}\selectfont
  \setlength{\tabcolsep}{1.25pt}
  \renewcommand{\arraystretch}{0.95}
  \begin{minipage}[t]{0.48\linewidth}
    \centering
    \textbf{(a)} Reward design\\[2pt]
    \begin{tabular}{l|cc|cc|c}
      \toprule
       & \textbf{MM-P} & \textbf{MM-R} & \textbf{Und.\,Avg} & \textbf{Gen.\,Avg} & \textbf{Time} \\
      \midrule
      \rev{X-Omni-RL} & 75.89 & 66.36 & \textcolor{ForestGreen}{52.66} & 42.50 & --- \\[0.25pt]
      \rowcolor{gray!15} Latent (Ours) & 76.93 & 69.42 & \textcolor{ForestGreen}{54.26} & 47.40 & 25\,h \\[0.25pt]
      SigLIP-2 & 76.92 & 67.84 & \textcolor{ForestGreen}{52.77} & 45.92 & 60\,h \\[0.25pt]
      DINOv3 & 75.78 & 66.93 & \textcolor{ForestGreen}{53.26} & 43.97 & 60\,h \\[0.25pt]
      MSE & 76.91 & 67.54 & \textcolor{ForestGreen}{53.43} & 42.29 & 58\,h \\[0.25pt]
      GPT-4.1 & 75.08 & 66.58 & \textcolor{ForestGreen}{52.38} & 44.15 & 68\,h \\[0.25pt]
      \bottomrule
    \end{tabular}
  \end{minipage}%
  \begin{minipage}[t]{0.48\linewidth}
    \centering
    \textbf{(b)} Sampling strategy\\[2pt]
    \begin{tabular}{l|cc|cc|c}
      \toprule
       & \textbf{MM-P} & \textbf{MM-R} & \textbf{Und.\,Avg} & \textbf{Gen.\,Avg} & \textbf{Time} \\
      \midrule
      Flat\,4 & 75.78 & 67.73 & \textcolor{ForestGreen}{53.14} & 43.60 & 22\,h \\
      $K{=}2,\,M{=}2$ & 76.47 & 68.45 & \textcolor{ForestGreen}{53.08} & 46.10 & 20\,h \\
      \midrule
      Flat\,16 & 77.03 & 67.83 & \textcolor{ForestGreen}{53.90} & 45.40 & 30\,h \\
      \rowcolor{gray!15} $K{=}4,\,M{=}4$ & 76.93 & 69.42 & \textcolor{ForestGreen}{54.26} & 47.40 & 25\,h \\
      \midrule
      $K{=}8,\,M{=}8$ & 76.57 & 68.36 & \textcolor{ForestGreen}{53.25} & 49.02 & 64\,h \\
      \bottomrule
    \end{tabular}
  \end{minipage}
  \caption{\rev{\textbf{(a)} Reward design: latent reward vs.\ alternative formulations. \textbf{(b)} Sampling strategy: $K/M$ sweep and flat GRPO comparison under matched budget. MM-P/R denote MMBench Perception/Reasoning. \textcolor{ForestGreen}{Und.\,Avg averages MMBench, MMVP, CV-Bench, and MathVerse; Gen.\,Avg averages WISE and GenEval++. Both are equally weighted means of percentage scores.}}}
  \label{tab:ablation_reward}
  \vspace{-3mm}
\end{table}

\subsection{Qualitative Case Study}
We use three qualitative probes to visualize instruction-grounded editing, WISE knowledge-grounded generation, and optimization dynamics under TreeGRPO.\linebreak Detailed WISE results and analysis are provided in Appendix~\ref{app:wise-results}.

\begin{figure}[!htb]
    \centering
    \includegraphics[width=0.90\linewidth]{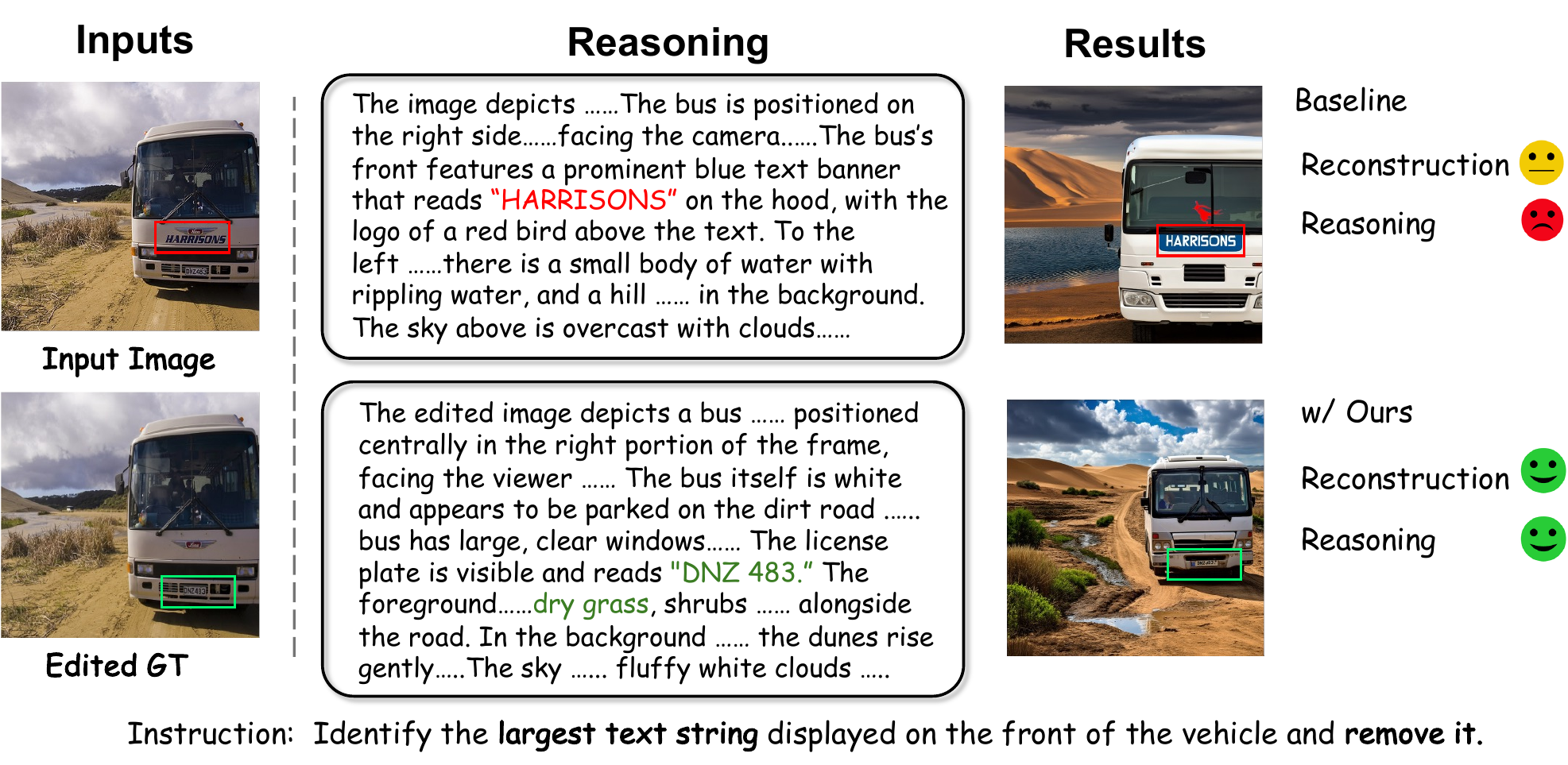}
    \caption{\rev{\textbf{Vehicle text-removal case.} We compare the intermediate target-state descriptions and final reconstructions under the same edit instruction. The baseline retains the target text, whereas ReasonUni removes the largest text string on the vehicle's front while preserving non-target details such as the license plate.}}
    \label{fig:rlreason}
\end{figure}

\paragraph{\textbf{Instruction-grounded editing.}}
\rev{In Figure~\ref{fig:rlreason}, the instruction is to remove the largest text string on the vehicle's front. X-Omni retains ``HARRISONS'' but mistakenly removes the license-plate text, whereas ReasonUni removes the intended text while preserving the plate. This example illustrates that our training improves not only fine-grained perception but also visual reasoning, enabling the model to infer the correct editing target from the instruction's spatial and comparative constraints. Both models receive the same prompt and edit instruction, so the difference stems from the trained weights rather than additional textual guidance. Additional examples and a reconstruction-only ablation (Appendices~\ref{app:qualitative}--\ref{app:extended-results}) further support the contribution of reasoning-augmented reconstruction to visual understanding.}

\begin{figure}[!t]
    \centering
    \includegraphics[width=0.90\linewidth]{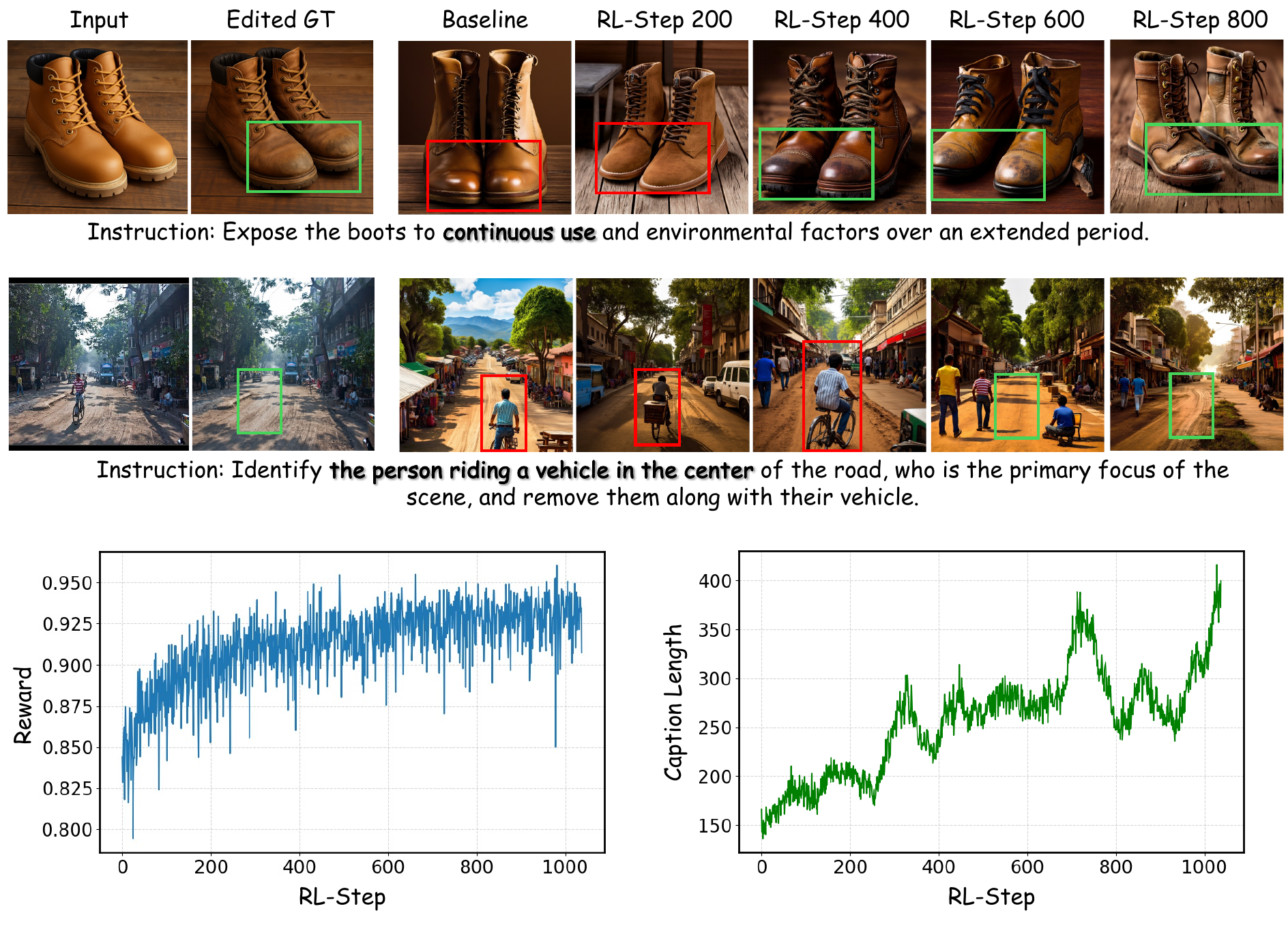}
    \caption{Case studies across RL training steps. Starting from a baseline model, ReasonUni progressively improves instruction-following edits as TreeGRPO training proceeds (Step 200 $\rightarrow$ 800). Top: the model increasingly captures long-term wear patterns on boots. \textcolor{ForestGreen}{Bottom: the model more accurately removes the rider in the center of the scene together with the vehicle.}}
    \label{fig:rlstep}
\end{figure}

\paragraph{\textbf{Optimization dynamics.}}
\textcolor{ForestGreen}{Figure~\ref{fig:rlstep} shows how editing behavior evolves during TreeGRPO training. Later checkpoints capture more visible wear on the boots and successfully remove the central rider and vehicle, indicating improved realization of the intended target state. The reconstruction reward rises rapidly early in training and gradually levels off, while target-state descriptions become longer overall. Together, these observations show changes in both intermediate descriptions and final reconstructions as training progresses, alongside improved reconstruction rewards.}

\begin{figure}[!htb]
    \centering
    \includegraphics[width=0.90\linewidth]{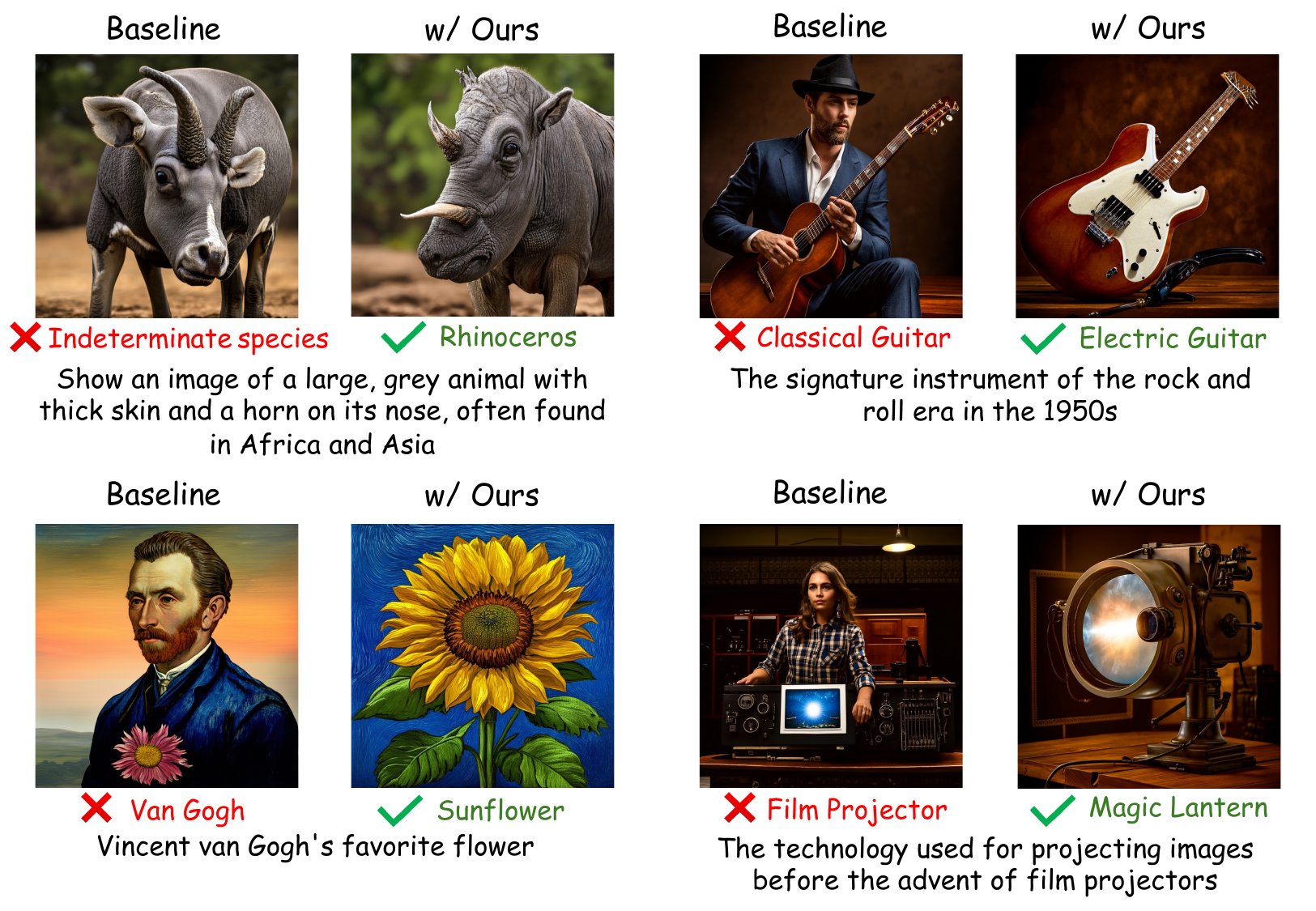}
    \caption{Qualitative results on WISE. \textcolor{ForestGreen}{For knowledge-focused prompts, ReasonUni generates more semantically accurate concepts than the baseline. The illustrated examples cover animal recognition, instrument categories, art-related knowledge, and historical technology.}}
    \label{fig:wisesample}
\end{figure}

\section{Related Work}

\paragraph{Architecture Design of Unified Multimodal Models.}
Unified Multimodal Models (UMMs) integrate understanding and generation within a single trainable framework~\cite{zhang2025unified}, with two dominant architectural paradigms.
\textbf{(1) Hybrid AR--Diffusion Architectures.}
\textcolor{ForestGreen}{These models combine autoregressive language modeling with diffusion-based visual generation. Show-o~\cite{xie2024show} uses discrete diffusion, while other approaches employ continuous diffusion or flow matching.} Transfusion~\cite{zhou2024transfusion} and related models~\cite{yan2025gpt,bagel,ma2024janusflow,shi2024llamafusion,xie2025show} share a backbone across modalities while using dedicated generative components, whereas another line connects a pre-trained MLLM to an external image generator through learnable adapters~\cite{pan2025metaquery,chen2025blip3,lin2025uniworld}.
\textbf{(2) Fully AR-based Approaches.}
\textcolor{ForestGreen}{These models generate both text and discrete image tokens autoregressively, while differing in how visual inputs are represented~\cite{team2024chameleon,qu2025tokenflow,wu2025janus,chen2025janus,liquid}.} Among fully autoregressive models, X-Omni~\cite{geng2025x} and Tar~\cite{han2025vision} further employ semantically aligned visual tokens, unifying visual understanding and generation within the same discrete representation space.

\paragraph{Post-Training in Unified Multimodal Models.}
Recent post-training methods for UMMs can be broadly grouped into two paradigms.
\textbf{(1) Post-training for Generation Quality.}
\begingroup\clubpenalties 3 10000 10000 0 These methods~\cite{chen2025blip3o,tong2025cot,jiang2025t2i,tian2025unigen,li2025uniworld,wei2025skywork} adapt reinforcement learning and related post-training techniques to improve generation and editing quality. T2I-R1~\cite{jiang2025t2i} further introduces reasoning over text prompts solely to improve image generation; in contrast, ReasonUni reasons over visual inputs and turns generation feedback into supervision for visual reasoning, jointly optimizing understanding and generation.\par\endgroup

\textbf{(2) Post-training for Understanding--Generation Alignment.}
This paradigm exploits the interaction between understanding and generation to improve both capabilities. Prior work explores self-generated supervision~\cite{han2026unicorn,jin2025srum,chen2026unified,niu2025does,tang2026endogenous,yang2025hermesflow,han2025turning} or joint reinforcement learning~\cite{jiang2025co,mao2025unirl}. More recently, reconstruction-based approaches~\cite{xie2025reca,yan2025uae} explicitly connect understanding and generation by conditioning image reconstruction on extracted semantics. \textbf{However, their supervision largely remains at the \textit{perception level}, providing limited explicit supervision for \textit{visual reasoning} and thus limiting generalization to tasks that require reasoning beyond perception.}

\section{Conclusion}
\textcolor{ForestGreen}{\texttt{\textbf{ReasonUni}} extends reconstructive post-training to visual reasoning, with \texttt{\textbf{TreeGRPO}} jointly optimizing understanding and generation through tree-structured reward comparisons. Across both initialization settings, ReasonUni achieves substantial gains on generation benchmarks alongside consistent improvements in visual understanding. Gains on all 12 evaluated understanding benchmarks, including improved logical reasoning on MMBench, further support its benefits beyond fine-grained perception. Together, these findings support reasoning-augmented reconstruction as an effective approach to using generation feedback for mutual gains in unified multimodal models.}

\clearpage  %
\setlength{\parfillskip}{0pt plus 1fil}

\bibliographystyle{iclr2027_conference}
\bibliography{main}
\clearpage
\appendix
\input{supp}

\end{document}

%% file: supp-preamble.tex
\usepackage{tcolorbox}
\tcbuselibrary{breakable}
\usepackage{float}
\definecolor{promptbg}{RGB}{245,245,245}
\definecolor{promptframe}{RGB}{220,220,220}
\newtcolorbox{promptbox}[1][]{
    breakable,
    before skip=10pt, after skip=10pt,
    colback=promptbg,
    colbacktitle=white,
    coltitle=black,
    colframe=black!75!black,
    boxrule=0.7pt,
    halign title=center,
    title=\textbf{#1}
}

%% file: authors.tex
\hypersetup{pdfauthor={Feize Wu, Zhiyuan Yan, Yuhuan Hong, Yuwei Niu, Zongjian Li, Zheyuan Liu, Bin Lin, Jinbo Yan, Jie Qin, Xudong Mao, Limeng Qiao, Guanglu Wan, Li Yuan}}
\author{%
  \parbox[t]{\dimexpr\textwidth-2\tabcolsep\relax}{%
    \centering
    {\fontsize{9}{11}\selectfont
    \mbox{\textbf{Feize Wu}$^{1,2,*}$, \textbf{Zhiyuan Yan}$^{1,*}$,
    \textbf{Yuhuan Hong}$^{3,*}$, \textbf{Yuwei Niu}$^{1}$,
    \textbf{Zongjian Li}$^{1}$, \textbf{Zheyuan Liu}$^{1}$}\\[2pt]
    \mbox{\textbf{Bin Lin}$^{1}$, \textbf{Jinbo Yan}$^{2}$, \textbf{Jie Qin}$^{2}$,
    \textbf{Xudong Mao}$^{4}$, \textbf{Limeng Qiao}$^{2}$,
    \textbf{Guanglu Wan}$^{2}$, \textbf{Li Yuan}$^{1,\dagger}$}\\[5pt]}
    \normalfont
    $^{1}$Peking University\qquad $^{2}$Meituan\\[2pt]
    $^{3}$Tsinghua University\qquad $^{4}$Sun Yat-sen University
  }%
}

\makeatletter
\gdef\@thanks{%
  \footnotetext[1]{Equal contribution. Contact: Feize Wu ({\hypersetup{pdfborder={0 0 0}}\href{mailto:wufeize26@stu.pku.edu.cn}{\textnormal{wufeize26@stu.pku.edu.cn}}}).}%
  \footnotetext[2]{Corresponding author.}%
}
\makeatother

%% file: supp.tex
\setlength{\parfillskip}{0pt plus .25\linewidth}
\raggedbottom
\makeatletter
\setlength{\@fptop}{0pt}
\setlength{\@fpsep}{16pt}
\makeatother
\section{Dataset Synthesis and Samples}
\label{app:data-synthesis}

\begin{figure}[!htbp]
    \centering
    \includegraphics[width=1\linewidth]{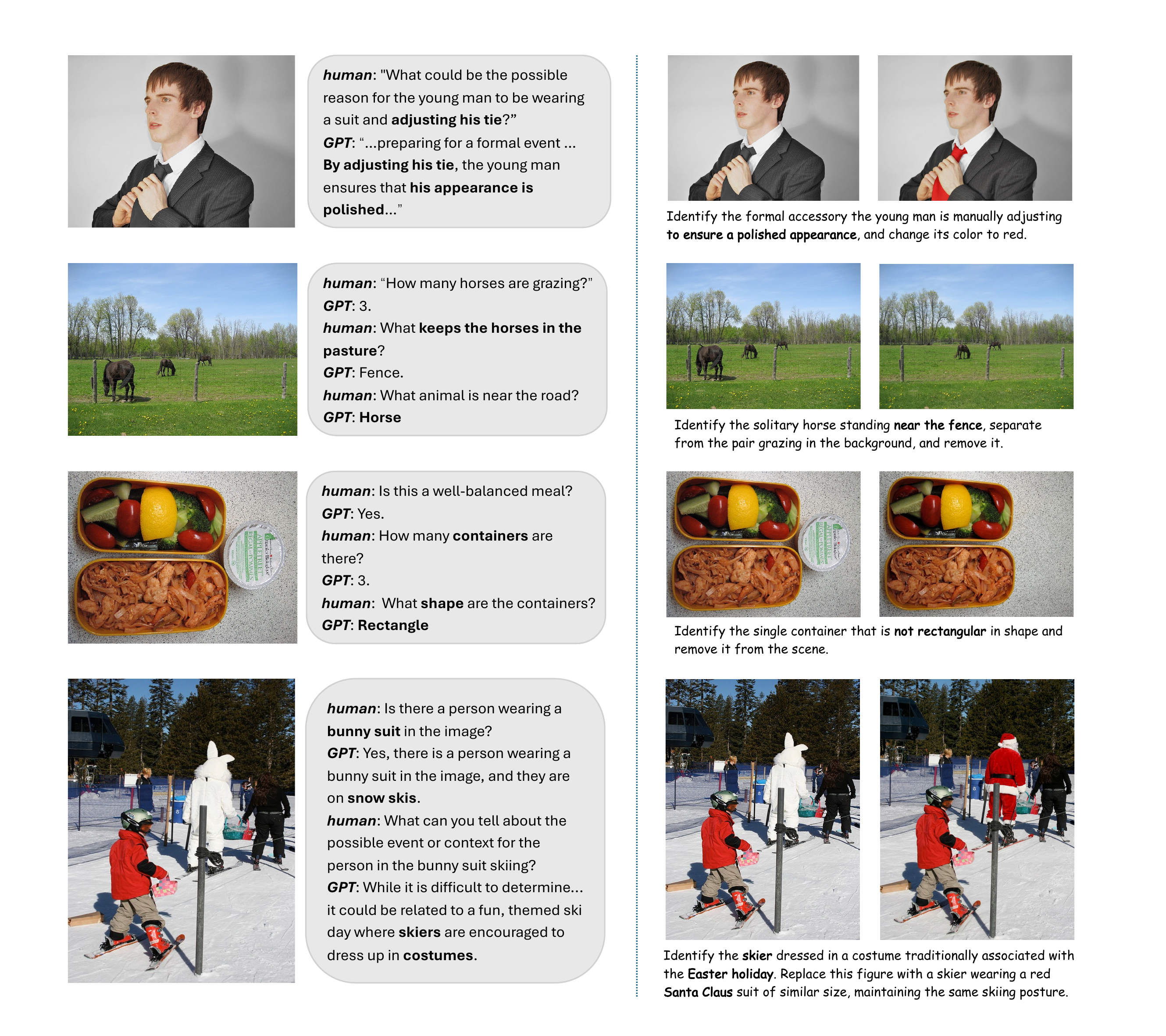}
    \caption{\textbf{Visualization of synthesized reasoning editing samples}. We construct the dataset by leveraging image-conversation pairs from LLaVA-NeXT-Data~\cite{liu2024llavanext}. \textbf{Left}: The source image and the original multi-turn dialogue serving as context. \textbf{Right}: The pair of source and edited images, with the synthesized instruction displayed below. The instruction is derived from the dialogue to guide the editing process. Each entry forms a complete tuple $(I_{src}, C_{hist}, T_{instr}, I_{edit})$, where the editing command is semantically grounded in the conversation history.
}
    \label{fig:reasoneidt-data-from-llava}
\end{figure}

To construct a dataset that demands high-level cognitive understanding for our reasoning-augmented reconstruction training, we developed a comprehensive data synthesis pipeline. Unlike traditional editing tasks that rely on explicit visual descriptions, our approach requires the model to identify target objects through logical analysis, spatial relationships, or common-sense inference. The overall process is divided into two primary stages, and visualizations of the synthesized data samples are provided in Figure \ref{fig:reasoneidt-data-from-llava}.

\subsection{Generation of Reasoning Instructions}

In the first stage, we leverage the rich multimodal contexts from LLaVA-NeXT-Data\cite{liu2024llavanext} as a foundational knowledge base. Specifically, we employ \textcolor{ForestGreen}{Gemini-3-Pro~\cite{google2025gemini3pro}} to deeply mine object relationships, functional attributes, spatial distributions, and physical laws from the original images. The detailed prompts are provided in Section~\ref{sec: prompt_templates}. This process simultaneously yields two critical components for each task.

\paragraph{Reasoning Instructions (The ``Logical Puzzle'').} Our pipeline synthesizes instructions by transforming contextual information into logical challenges. The instructions are designed to identify target objects through their functional or relational context within the scene rather than direct visual descriptions (e.g., ``the red cup'') (e.g., ``Identify the vessel on the desk used for holding hot beverages and change its color''). To ensure comprehensive cognitive coverage, we defined twelve distinct reasoning categories, including spatial relations, quantitative comparison, causal prediction, and physical attributes, spanning from basic perception to advanced cognitive vision.

\paragraph{Deterministic Outcomes and Reference Descriptions.} Our pipeline aims to generate instructions with specific, relatively deterministic outcomes to reduce ambiguity in the reconstruction target. Concurrently with the implicit instruction, the LLM generates a reference effect, i.e., a descriptive, explicit text of the post-edit state. This reference text is used to guide the generation of the ground-truth image in the subsequent stage.

\subsection{Instruction-Guided Image Editing and Execution}

Once the self-contained reasoning instructions and their corresponding reference descriptions are generated, they are fed into a Nano Banana to produce the ground-truth edited images (the image generation prompt is detailed in Section~\ref{sec: prompt_templates}).

\paragraph{Precise Visual Constraint.} During the image synthesis phase, the explicit reference description is provided alongside the reasoning instruction to the generative model. The reference description specifies the intended post-edit state and guides\linebreak the synthesis of visual details such as textures, lighting, and materials.

\paragraph{Contextual Coherence.} When performing operations such as removal, replacement, or position swapping, the synthesis process aims to preserve plausible background completion and consistent lighting.
By employing this two-stage methodology, we successfully generate high-quality data tuples consisting of the original image, reasoning instruction, reference effect and the edited result, which provide the diverse, challenging reasoning instructions and accurate ground-truth edits essential for our Reasoning-Augmented Reconstruction. \textcolor{ForestGreen}{Furthermore, we will open-source the synthesized dataset of 30,000 samples.} This initiative will not only facilitate the open-source community in adopting our method, but also provide a robust training and evaluation resource for the broader field of reasoning-guided editing.

\section{Extended Implementation Details}
\label{app:implementation}

\subsection{Baselines and Ablation Study Implementation Details}
\label{app:baseline-details}

\paragraph{Base Model and Main Experiments.} Our proposed approach operates entirely within the post-training paradigm. The main experiments are conducted based on the X-Omni model architecture\cite{geng2025x}. \textcolor{ForestGreen}{X-Omni is a pure AR UMM that represents images with semantically aligned SigLIP-VQ visual tokens.} \textcolor{blue}{We initialize from the official X-Omni-SFT and X-Omni-RL checkpoints, corresponding to the SFT and RL settings defined in the main paper.}

\paragraph{\textcolor{ForestGreen}{Training Data and Configuration.}}
{\color{ForestGreen}We train on a sampled mixture of editing data from UniREdit-Data-100K~\cite{han2025unireditbench}, RISE~\cite{zhao2026envisioning}, and KRIS~\cite{wu2026kris}, together with synthetic reasoning-driven editing data. The latter are adapted from ReasonSeg~\cite{lai2024lisa}, LLaVA-NeXT-Data~\cite{liu2024llavanext}, and TextVQA~\cite{singh2019towards} using Gemini-3-Pro~\cite{google2025gemini3pro}. We train for one epoch on 8 H100 GPUs with a peak learning rate of $1 \times 10^{-5}$, per-device batch size 1, two-step gradient accumulation, bfloat16, ZeRO Stage 1, gradient checkpointing, and LoRA ($r{=}8$, $\alpha{=}32$). Prompts are truncated to 512 tokens, both KL coefficients are 0.003, and the gradient norm is clipped at 1.0. TreeGRPO uses $K{=}4$ reasoning paths and $M{=}4$ images per path, yielding 16 images per input.}

\paragraph{Baseline Implementation.} For baseline comparisons and ablation studies, we incorporate UAE\cite{yan2025uae}, an alternative post-training method. Because the official UAE release does not include an implementation for the X-Omni architecture, we ported and re-implemented UAE on X-Omni for this study. \textcolor{ForestGreen}{Unmarked UAE results use its official training data. The matched-data control (UAE$^\dagger$ in Table~\ref{tab:main_results}) uses the same initialization checkpoint, training samples, and number of training steps as ReasonUni. We use the same optimization hyperparameters for UAE and ReasonUni.}

\paragraph{\textcolor{ForestGreen}{Reconstruction-only Ablation.}}
\textcolor{ForestGreen}{The ``+TreeGRPO'' rows in Table~\ref{tab:performance_comparison}, labeled ``Ours w/o Reasoning'' in Figures~\ref{fig:geneval++-sample-supp} and~\ref{fig:wisesample-supp}, retain TreeGRPO but use only source-image reconstruction. The understanding module describes the source image without an editing instruction, and the generation module reconstructs that same image from the description, rather than an edited target state.}

\subsection{Evaluation Details}

\paragraph{Visual Understanding Evaluation.} To assess visual understanding capabilities, we utilize the VLMEvalKit framework\cite{duan2024vlmevalkit}. Throughout the evaluation process, GPT-4o\cite{gpt4o} is employed for all operations requiring external API calls. The inference implementation and system prompts strictly follow the official X-Omni codebase. The inference temperature for visual understanding tasks is set to 0.

\paragraph{Image Generation Evaluation.} For image generation tasks, we similarly adopt the official X-Omni inference code and parameter configurations. The generation resolution is fixed at $1152\times1152$. The core sampling parameters are configured with a temperature of 1.0, a min-$p$ value of 0.03, a top-$p$ value of 1.0, and a classifier-free guidance (CFG) scale of 1.0. For specific generation benchmarks, i.e., DPG-Bench\cite{hu2024ella}, WISE\cite{wisebench}, and GenEval$++$\cite{ye2025echo}, we follow their official evaluation pipelines and use their designated models for scoring.

\clearpage
\section{Qualitative Analysis and Visualizations}
\label{app:qualitative}

\begin{figure}[!htbp]
    \centering
    \includegraphics[width=0.9\linewidth]{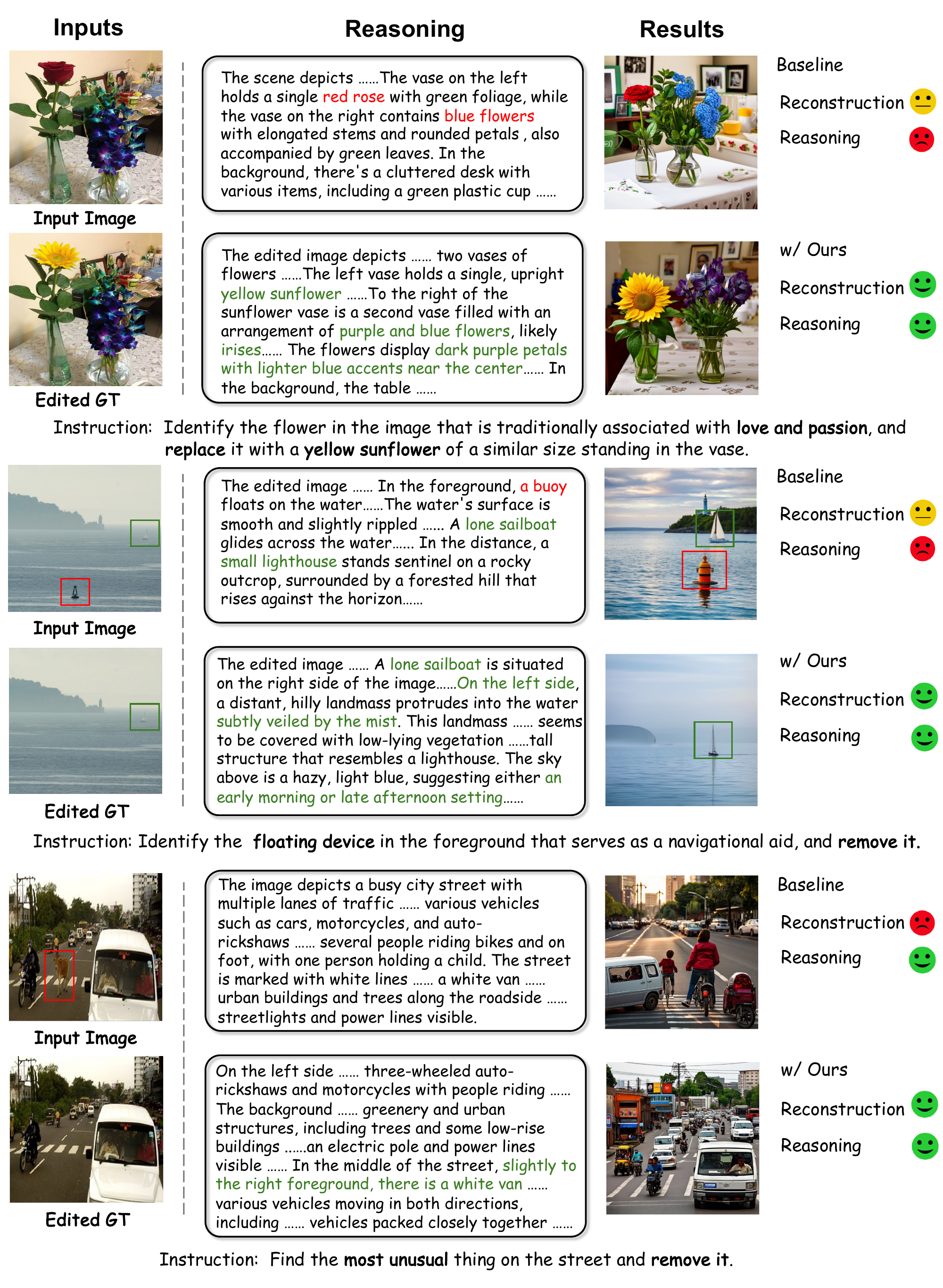}
    \caption{\textcolor{blue}{\textbf{Additional instruction-based editing cases.} We compare baseline and ReasonUni target-state descriptions and reconstructions across floral, maritime, and street scenes. Across the three cases, ReasonUni better matches the displayed edited references while introducing fewer unrelated changes.}}
    \label{fig:rlreason-supp}
\end{figure}

\begin{figure}[!htbp]
    \centering
    \includegraphics[width=0.9\linewidth]{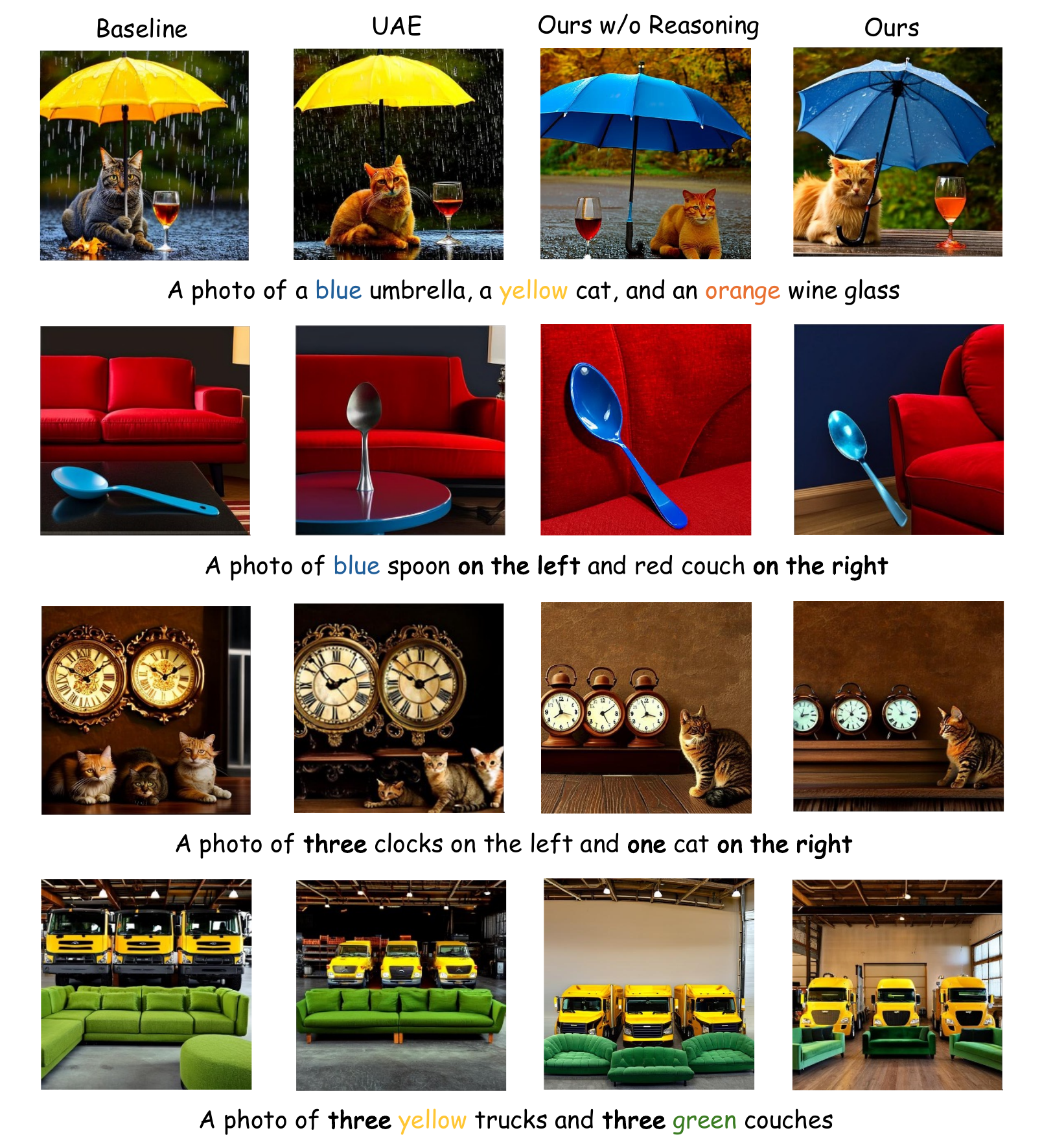}
    \caption{\textbf{Qualitative results on GenEval++}. Compared with the baselines, ReasonUni better preserves compositional attributes, object counts, and spatial relations.}
    \label{fig:geneval++-sample-supp}
\end{figure}

We provide additional qualitative evidence to complement the quantitative results from multiple perspectives. \textcolor{blue}{Figure~\ref{fig:rlreason-supp} presents comparisons across floral, maritime, and street-scene cases. Across these cases, ReasonUni's target-state descriptions better match the displayed edited references, and its reconstructions preserve non-target content more consistently.} Figure~\ref{fig:geneval++-sample-supp} further highlights stronger compositional prompt-following ability, where our full model preserves object attributes, counts, and spatial bindings more consistently than the baseline variants.

\begin{figure}[!htbp]
    \centering
    \includegraphics[width=0.9\linewidth]{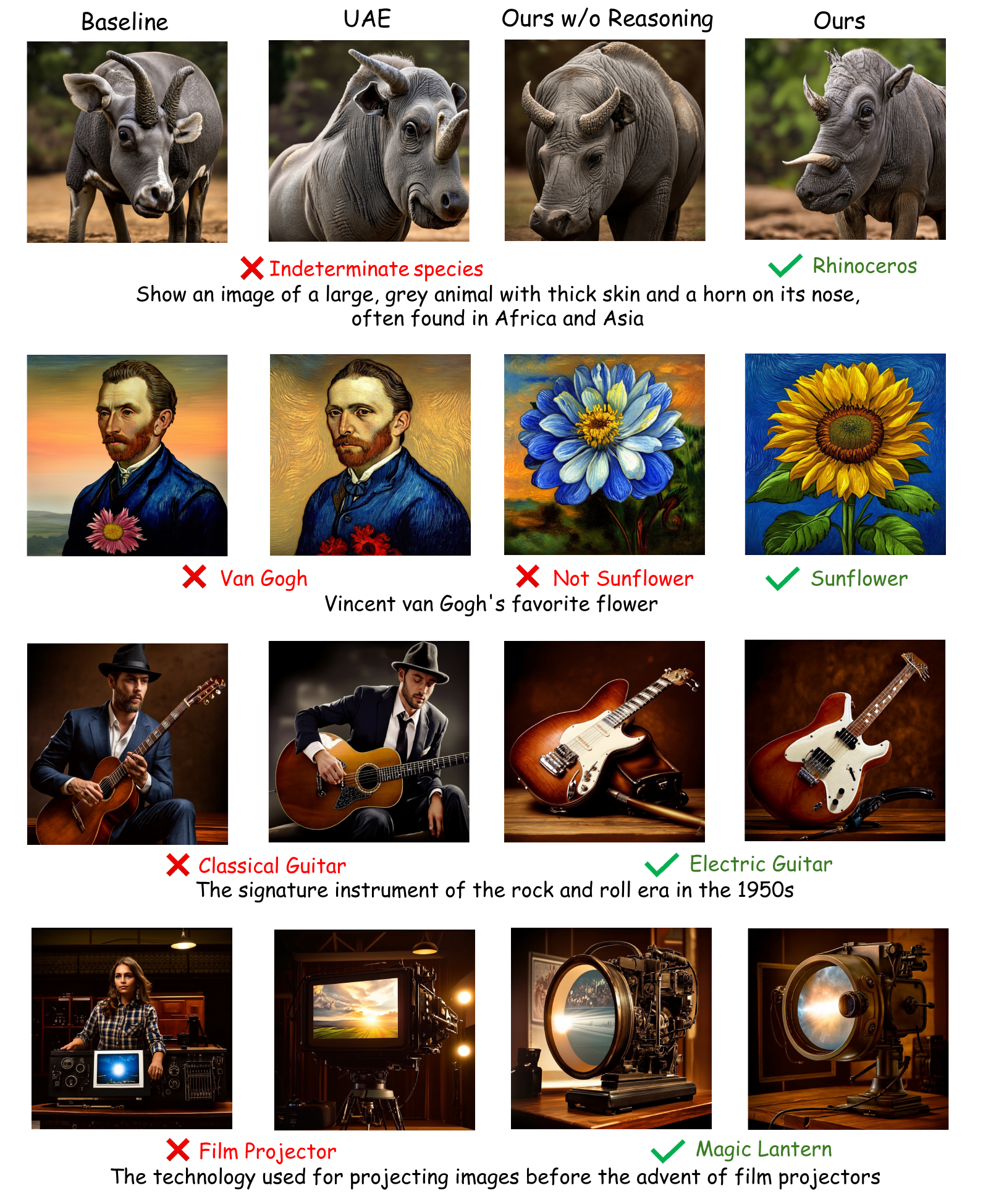}
    \caption{\textbf{Qualitative results on WISE.} For knowledge-focused prompts, ReasonUni generates more semantically accurate concepts than the baseline, including animal recognition, instrument category, art-related knowledge, and historical technology.}
    \label{fig:wisesample-supp}
\end{figure}

\clearpage
\section{Extended Results and Ablation Studies}
\label{app:extended-results}

This section provides additional quantitative evidence for both effectiveness and mechanism analysis of our method.
A detailed fine-grained comparison on DPG-Bench is presented in Table~\ref{tab:dpgbench_subtask_comparison}, where our method achieves stronger prompt-following performance than the X-Omni baseline on overall score and most sub-metrics.
\textcolor{ForestGreen}{Additionally, Table~\ref{tab:performance_comparison} compares reconstruction-only TreeGRPO with the full reasoning-augmented framework across the SFT and RL settings, evaluating both understanding and generation.}

The subsequent multimodal understanding comparison further breaks down performance by capability domain and indicates that the most notable gains come from vision-centric evaluation while maintaining stable hallucination and OCR behavior. Finally, Table~\ref{tab:comparison} reports detailed generation results on DPGBench, WISE, and GenEval$++$, highlighting improved compositional faithfulness in relation/attribute following, knowledge-grounded generation, and position/count control.

\begin{table}[!htbp]
\caption{\textbf{Fine-grained evaluation on DPGBench\cite{hu2024ella}.} We report the detailed breakdown of performance on Global, Entity, Attribute, Relation, and Other metrics. Compared to the baseline X-Omni, our method demonstrates superior capability in following fine-grained text prompts.}
\label{tab:dpgbench_subtask_comparison}
\centering
\scriptsize
 \resizebox{\textwidth}{!}{
\renewcommand{\arraystretch}{1.2}
\ %
\begin{tabular}{lcccccc}
\toprule
\textbf{Model} & \textbf{Global} & \textbf{Entity} & \textbf{Attribute} & \textbf{Relation} & \textbf{Other} & \textbf{Overall} \\
\midrule

Emu3\cite{wang2024emu3} & 85.21 & 86.68 & 86.84 & 90.22 & 83.15 & 80.60 \\
Janus-Pro\cite{chen2025janus} & 86.90 & 88.90 & 89.40 & 89.32 & 89.48 & 84.19 \\
MetaQuery\cite{pan2025metaquery} & - & - & - & - & - & 82.05 \\
BLIP3-o\cite{chen2025blip3} & - & - & - & - & - & 81.60 \\
UniWorld-V1\cite{lin2025uniworld} & 83.64 & 88.39 & 88.44 & 89.27 & 87.22 & 81.38 \\
Ovis-U1\cite{wang2025ovis} & 82.37 & 90.08 & 88.68 & 93.35 & 85.20 & 83.72 \\
Mogao\cite{liao2025mogao} & 82.37 & 90.03 & 88.26 & 93.18 & 85.40 & 84.33 \\
BAGEL\cite{bagel} & 88.94 & 90.37 & 91.29 & 90.82 & 88.67 & 85.07 \\
OmniGen2\cite{wu2025omnigen2} & 88.81 & 88.83 & 90.18 & 89.37 & 90.27 & 83.57 \\
\midrule

X-Omni\cite{geng2025x} & 90.10 & 87.75 & 89.21 & 88.54 & 88.09 & 82.29 \\
\rowcolor{gray!15}+Ours& {87.28} & {89.79} & {90.29} &{90.70} & {88.12} & {83.58} \\
\bottomrule
\end{tabular}
}
\vspace{1mm}

\end{table}

\begin{table*}[!htbp]
\caption{\textbf{Ablation study of proposed components.} \textcolor{blue}{We evaluate the effectiveness of TreeGRPO and reasoning augmentation across the SFT and RL settings.}}
\label{tab:performance_comparison}
  \centering
  \scriptsize
  \renewcommand{\arraystretch}{1.2}
  \begin{tabular}{l cccc cccc}
    \toprule
    \multirow{2}{*}{\textbf{Method}} & \multicolumn{4}{c}{\textbf{Understanding}} & \multicolumn{3}{c}{\textbf{Generation}} \\
    \cmidrule(lr){2-5} \cmidrule(lr){6-8}
    & MMBench & MMVP & CVBench & MathVerse & DPGBench & WISE & Geneval++ \\
    \midrule

    \textcolor{blue}{X-Omni-SFT} & 74.4 & 48.67 & 67.98 & 17.01 & 65.25 & 24 & 7.5 \\
    +TreeGRPO & 75.51 & 50.00 & \textbf{69.20} & 16.75 & 77.24 & 34 & 20.36 \\
    \rowcolor{gray!15} +TreeGRPO +Reasoning & \textbf{75.60} & \textbf{52.67} & 68.85  & \textbf{17.77} & \textbf{78.59} & \textbf{37.00} & \textbf{23.21} \\

    \midrule

    \textcolor{blue}{X-Omni-RL} & 74.31 & 50.67 & 68.26 & 17.39 & 82.29 & 45 & 40 \\
    +TreeGRPO & 75.86 & 50.67 & 67.91 & 17.36 & 82.98 & 45 & 43.21 \\
    \rowcolor{gray!15} +TreeGRPO +Reasoning & \textbf{75.96} & \textbf{52.67} & \textbf{70.01} & \textbf{18.40} & \textbf{83.58} & \textbf{48} & \textbf{46.79} \\

    \bottomrule
  \end{tabular}
  \vspace{1mm}

\end{table*}

\begin{table}[!htbp]
\caption{\textbf{Comparison with SOTA unified models.} We conduct a comprehensive evaluation, including \textbf{General} (MMBench\cite{liu2023mmbench}, SEED-Bench\cite{li2023seed}, MMStar\cite{chen2024mmstar}, MMT\cite{mmt}), \textbf{Vision-Centric} (VStar\cite{wu2024vstar}, MMVP\cite{tong2024mmvp}, CV-Bench\cite{tong2024cvbench}), \textbf{Hallucination} (POPE\cite{li2023pope}, HallusionBench\cite{guan2024hallusionbench}), \textbf{OCR} (OCRBench\cite{liu2024ocrbench}), \textbf{Knowledge} (MMMU)\cite{yue2024mmmu}, and \textbf{Math} (MathVerse\cite{zhang2024mathverse}). ``--'' denotes results not reported in original papers.}

  \centering
  \resizebox{\textwidth}{!}{%
    \renewcommand{\arraystretch}{1.2}
    \begin{tabular}{l cccc ccc cc c c c}
    \toprule
    \multirow{2}{*}{\textbf{Method}} & \multicolumn{4}{c}{\textbf{General}} & \multicolumn{3}{c}{\textbf{Vision-Centric}} & \multicolumn{2}{c}{\textbf{Hallucination}} & \textbf{OCR} & \textbf{Knowledge} & \textbf{Math} \\
    \cmidrule(lr){2-5} \cmidrule(lr){6-8} \cmidrule(lr){9-10} \cmidrule(lr){11-11} \cmidrule(lr){12-12} \cmidrule(lr){13-13}
    & MMB & SEED & MMStar & MMT & VStar & MMVP & CVB. & POPE & Hallu. & OCRB. & MMMU & MathV. \\
    \midrule
    Show-o\cite{xie2024show}    & 47.9 & --    & --    & --    & --    & 50.0  & --    & 73.8  & --    & --    & 25.1  & -- \\
    Emu3\cite{wang2024emu3}      & 58.5  & 68.2  & --    & --    & 51.8  & 36.6  & --    & 85.2  & --    & 68.7  & 31.6  & -- \\
    VILA-U\cite{wu2024vila}    & 66.6  & 57.1  & --    & --    & 38.7  & 22.0  & --    & 85.8  & --    & --    & 32.2  & -- \\
    Janus Pro\cite{chen2025janus} & 79.2  & 72.1  & --    & --    & --    & --    & --    & 87.4  & --    & 59.5  & 41.0  & -- \\
    MetaMorph\cite{tong2025metamorph} & 75.2  & 71.8  & --    & --    & 44.0  & 48.3  & --    & --    & --    & --    & 41.8  & -- \\
    Mogao\cite{liao2025mogao}     & 75.0  & 74.6  & --    & --    & --    & --    & --    & 88.9  & --    & --    & 44.2  & -- \\
    Bagel\cite{bagel}     & 79.2  & --    & --    & --    & --    & 54.7  & --    & --    & --    & --    & 43.2  & -- \\
    \midrule
    Tar\cite{han2025vision}    & 69.67 & 70.62 & 44.33 & 50.53 & 39.27 & 46.00 & 61.93 & 85.14 & 54.47 & 34.3  & 38.00    & 16.12 \\
    \rowcolor{gray!15}+Ours     & 70.96 & 71.02 & 44.67 & 51.61 & 42.41 & 48.00 & 62.79 & 85.39 & 54.57 & 34.1  & 37.78    & 16.48 \\
    \midrule
    Gain ($\uparrow$) & \textcolor{GainGreen}{+1.29} & \textcolor{GainGreen}{+0.40} & \textcolor{GainGreen}{+0.34} & \textcolor{GainGreen}{+1.08} & \textcolor{GainGreen}{+3.14} & \textcolor{GainGreen}{+2.00} & \textcolor{GainGreen}{+0.86} & \textcolor{GainGreen}{+0.25} & \textcolor{GainGreen}{+0.10} & \textcolor{Gray}{-0.20} & \textcolor{Gray}{-0.22} & \textcolor{GainGreen}{+0.36} \\
    \bottomrule
    \end{tabular}
  }
  \vspace{1mm}

\end{table}

\begin{table}[!htbp]
\caption{\textbf{Comparisons on generation benchmarks.} We report performance on DPGBench\cite{hu2024ella}, WISE\cite{wisebench}, and GenEval++\cite{ye2025echo}. Compared to the baseline and other unified models, our approach demonstrates favorable performance, showing consistent gains in attribute binding, world knowledge, and complex query alignment.}
\label{tab:comparison}
  \centering
  \resizebox{\textwidth}{!}{%
    \renewcommand{\arraystretch}{1.2}

    \begin{tabular}{l cccc cccc cccc}
    \toprule
    \multirow{2}{*}{\textbf{Method}} & \multicolumn{4}{c}{\textbf{DPGBench}} & \multicolumn{4}{c}{\textbf{WISE}} & \multicolumn{4}{c}{\textbf{GenEval++}} \\
    \cmidrule(lr){2-5} \cmidrule(lr){6-9} \cmidrule(lr){10-13}

    & Rel. & Attr. & Other & Overall & Cult. & Time  & Phys. & Overall & Pos/Cnt & Pos/Size & Cnt & Overall \\
    \midrule
    Harmon~\cite{wu2025harmon} & - & - & - & - & 38 & 48 & 44 & 41 & - & - & - & - \\
    Show-o~\cite{xie2024show} & - & - & - & - & 28 & 40 & 46 & 35 & - & - & - & - \\
    Emu3\cite{wang2024emu3} & 90.22 & 86.84 & 83.15 & 80.60 & - & - & - & - & - & - & - & - \\
    Janus-Pro\cite{chen2025janus} & 89.32 & 89.40 & 89.48 & 84.19 & 30 & 37 & 42 & 35 & 7.5 & 35.0 & 30.0 & 24.60 \\
    MetaQuery\cite{pan2025metaquery} & - & - & - & 82.05 & 56 & 55 & 63 & 55 & - & - & - & - \\

    BLIP3-o\cite{chen2025blip3} & - & - & - & 81.60 & - & - & - & - & 12.5 & 57.5 & 25.0 & 30.70 \\
    UniWorld-V1\cite{lin2025uniworld} & 89.27 & 88.44 & 87.22 & 81.38 & 53 & 55 & 59 & 55 & - & - & - & - \\
    Ovis-U1\cite{wang2025ovis} & 93.35 & 88.68 & 85.20 & 83.72 & - & - & - & - & - & - & - & - \\
    Mogao\cite{liao2025mogao} & 93.18 & 88.26 & 85.40 & 84.33 & - & - & - & - & - & - & - & - \\

    OmniGen2\cite{wu2025omnigen2} & 89.37 & 90.18 & 90.27 & 83.57 & 42 & 52 & 50 & 47 & 12.5 & 25.0 & 42.5 & 32.50 \\
    T2I-R1\cite{jiang2025t2i} & - & - & - & - & - & - & - & - & 7.5 & 25.0 & 32.5 & 31.10 \\
    \midrule
    Tar\cite{han2025vision}    & 88.59 & 86.99 & 83.99 & 78.99 & 46 & 49  & 45 & 46 & 10 & 50 & 25 & 30.36 \\

    \rowcolor{gray!15}+Ours  & 89.38 & 87.38 & 86.31 & 79.38 & 49 & 50  & 47 & 47 & 15 & 60 & 35 & 31.79 \\
    \midrule

    Gain ($\uparrow$) & \textcolor{GainGreen}{+0.79} & \textcolor{GainGreen}{+0.39} & \textcolor{GainGreen}{+2.32} & \textcolor{GainGreen}{+0.39} & \textcolor{GainGreen}{+3.00} & \textcolor{GainGreen}{+1.00}  & \textcolor{GainGreen}{+2.00} & \textcolor{GainGreen}{+1.00} & \textcolor{GainGreen}{+5.00} & \textcolor{GainGreen}{+10.0} & \textcolor{GainGreen}{+10.0} & \textcolor{GainGreen}{+1.43} \\
    \bottomrule
    \end{tabular}
  }
  \vspace{1mm}

\end{table}

\subsection{WISE Results and Analysis}
\label{app:wise-results}

\begin{table}[!htbp]
\caption{\textbf{Performance comparison on WISE.} WISE assesses knowledge-grounded generation across six domains: Cultural, Time, Space, Biology, Physics, and Chemistry. ReasonUni improves over X-Omni-RL in every domain and achieves an overall score of 48.}
\label{tab:wise_subtask_comparison}
  \centering
  \resizebox{\textwidth}{!}{%
    \renewcommand{\arraystretch}{1.25}
    \setlength{\tabcolsep}{5pt}
    \begin{tabular}{l cccccc c}
    \toprule
    \textbf{Model} & \textbf{Cultural} & \textbf{Time} & \textbf{Space} & \textbf{Biology} & \textbf{Physics} & \textbf{Chemistry} & \textbf{Overall} \\
    \midrule
    Harmon       & 38 & 48 & 52 & 37 & 44 & 29 & 41 \\
    Show-o       & 28 & 40 & 48 & 30 & 46 & 30 & 35 \\
    Janus Pro    & 30 & 37 & 49 & 36 & 42 & 26 & 35 \\
    MetaQuery-XL & 56 & 55 & 62 & 49 & 63 & 41 & 55 \\
    UniWorld-V1  & 53 & 55 & 73 & 45 & 59 & 41 & 55 \\
    OmniGen2     & 42 & 52 & 64 & 43 & 50 & 34 & 47 \\
    \midrule
    X-Omni-RL    & 47 & 47 & 55 & 38 & 48 & 28 & 45 \\
    \rowcolor{gray!15}+ Ours & 50 & 49 & 60 & 41 & 52 & 32 & 48 \\
    \bottomrule
    \end{tabular}%
  }
  \vspace{1mm}

\end{table}

Compared with the baseline, ReasonUni generates semantically correct concepts across multiple domains, including animal taxonomy (rhinoceros), instrument category (electric guitar), art-related knowledge (sunflower), and historical technology (magic lantern). WISE overall improves from 24 to 37 \textcolor{blue}{in the SFT setting} and from 45 to 48 \textcolor{blue}{in the RL setting}; \textcolor{blue}{in the RL setting}, all six domains improve over X-Omni-RL (Cultural +3, Time +2, Space +5, Biology +3, Physics +4, Chemistry +4), as shown in Tables~\ref{tab:performance_comparison} and~\ref{tab:wise_subtask_comparison}. \textcolor{ForestGreen}{Figure~\ref{fig:wisesample-supp} further illustrates ReasonUni's improved ability to generate the intended concepts rather than plausible but incorrect substitutes.} The largest gains on Space, Physics, and Chemistry indicate that reasoning-augmented reconstruction improves not only concept recall but also knowledge-grounded compositional generation.

\clearpage
\setlength{\parfillskip}{0pt plus 1fil}
\section{Prompt Templates}
\label{sec: prompt_templates}

\begin{promptbox}[Template for Reasoning-Caption Generation]

You are a helpful assistant. Based on the input image provided above and the complex editing instruction that requires reasoning, imagine the edited image and directly generate a textual description of the result.
\\ \\
Requirements for the description:\\
1. Describe the scene layout.  \\
2. Include details of people, animals, or objects.  \\
3. Mention environmental details.  \\
4. If there is text in the image, transcribe it inside double quotes "".  \\

Reasoning Instruction: \{instruction\} \\
\\
Now, carefully analyze the input image and the instruction, reason about the changes, and directly output the full description of the edited result. Do not output the reasoning process.

\end{promptbox}

\begin{promptbox}[Prompt for Edited GT Generation in Data Pipeline]
You are an expert image editor and generative AI system specialized in creating realistic visual transformations and edits. Your task is to generate images that depict the exact state described in the reference description, following the user's instruction precisely. \\
\\
Key Principles: \\
1. ACCURACY: Generate images that precisely match the reference description provided \\
2. REALISM: Ensure all visual details are photorealistic and plausible \\
3. COHERENCE: The generated image should logically follow from the instruction provided \\
4. DETAIL: Include specific visual details mentioned in the reference description \\
5. QUALITY: Produce high-quality, clear, and well-composed images \\
\\
Guidelines: \\
- Focus on the exact state after the transformation/operation described \\
- Include all relevant visual details from the reference description \\
- Maintain appropriate composition and lighting \\
- Generate complete, finished images (not sketches or partial views) \\
- Ensure colors, textures, and materials are rendered authentically''

\end{promptbox}

\clearpage
\begin{promptbox}[Prompt for Instruction Generation in Data Pipeline]
\small

You are an expert in image reasoning and editing.Your task is to analyze the provided image and the conversation history to generate a self-contained ``Implicit Reasoning Instruction'' for image editing. \\
\\
The goal is to extract a reasoning challenge from the conversation and formulate an editing instruction where the target object is NOT explicitly named, but must be inferred through reasoning (the ``implicit'' part). The instruction should act like a puzzle that requires understanding the image's logic, spatial relations, or common sense to solve before editing. \\
\\
\textbf{Reasoning Types:} \\
- quantity (counting, specific numbers) \\
- size (largest, smallest, taller) \\
- distance (closest, furthest) \\
- depth (foreground, background) \\
- attribute (color, material, old/new, function) \\
- relation (next to, on top of, holding, above, below) \\
- common sense (weather, gravity, biology, physics) \\
- knowledge (specific entities, brands, historical facts, object purposes) \\
- temporal (before/after, sequence of events, duration) \\
- causal (cause-effect, if-then, consequence) \\
- perspective (point of view, reflection, mirror) \\
- composition (parts of a whole, arrangement) \\
\\
\textbf{Edit Types:} \\
- change color \\
- removal -- Remove completely, leaving plausible background) \\
- replace with similar-sized object -- MUST explicitly and detailed describe what new object should replace the target, ensuring it is of similar size and fits the context \\
- future prediction -- MUST include time frame (e.g., ``2 seconds later'', ``after the action is completed'') \\
- swap positions \\
\\
\textbf{Requirements:} \\
1. Implicit Target Identification: The \texttt{instruction} MUST NOT explicitly describe the visual appearance of the target object if it can be avoided. Instead, use the logical relationships, questions, or functional descriptions from the conversation to point to the object. \\
- BAD: ``Edit the image to change the color of the brown slatted window blinds located directly above the stove to white.'' (Too explicit) \\
- GOOD: ``Identify the object on the wall above the microwave, and change its color to white.'' (Requires spatial reasoning) \\

- BAD: ``Remove the ceramic bowl containing a spoon located in the foreground.'' \\
- GOOD: ``Identify the item that suggests the person recently ate breakfast, and remove it from the desk.'' (Requires common sense/knowledge reasoning) \\
2. \textbf{Self-Contained:} The \texttt{instruction} MUST NOT mention ``the conversation'', ``the user'', ``as discussed'', or ``the object mentioned''. It must stand alone as a valid prompt for an image editing model. \\
3. \textbf{Explicit Action:} The \texttt{instruction} MUST explicitly command the model to generate the result or perform an edit (e.g., ``Draw the result of...'', ``Generate an image where...'', ``Edit the image to...'', ``Identify X and remove it''). \\
4. \textbf{Plausibility:} The edit must be physically plausible and maintain the coherence of the original image. \\
5. \textbf{Determinism:} The expected editing result MUST be relatively deterministic and specific. It should not be vague or open to multiple vastly different interpretations. The instruction and the \texttt{edited\_effect} should clearly define what the final image looks like. \\
\\
\tcbbreak
\textbf{Few-Shot Examples:} \\
\\
\textbf{Example 1 (Relation + Change Color)} \\
* \textbf{Context:} User asks ``What is on the wall above the microwave?'' and GPT answers ``Cabinet''. \\
* \textbf{Output:} \\
\texttt{\{} \\
\texttt{\quad "reasoning\_type": "relation",} \\
\texttt{\quad "edit\_target": "the cabinet above the microwave",} \\
\texttt{\quad "edit\_type": "change color",} \\
\texttt{\quad "instruction": "Identify the object on the wall above the microwave, and change its color to white.",} \\
\texttt{\quad "edited\_effect": "The cabinet located above the microwave is now white."} \\
\texttt{\}} \\
\\
\textbf{Example 2 (Common Sense + Future Prediction)} \\
* \textbf{Context:} User asks ``Where does the man want the ball to go? A. behind him B. in front C. in pocket D. in hand'' and GPT answers ``B''. The image shows a man preparing to serve a tennis ball. \\
* \textbf{Output:} \\
\texttt{\{} \\
\texttt{\quad "reasoning\_type": "common sense",} \\
\texttt{\quad "edit\_target": "the tennis ball",} \\

\texttt{\quad "edit\_type": "future prediction",} \\
\texttt{\quad "instruction": "Based on where the man wants the ball to go, draw the result happening 2 seconds after the current scene.",} \\

\texttt{\quad "edited\_effect": "The man has completed his serve, and the ball is no longer in his hand or in the air near him, but has been hit forward."} \\
\texttt{\}} \\

\textbf{Example 3 (Quantity + Replacement)} \\
* \textbf{Context:} User asks how many apples are on the desk. \\
* \textbf{Output:} \\
\texttt{\{} \\
\texttt{\quad "reasoning\_type": "quantity",} \\
\texttt{\quad "edit\_target": "the four apples",} \\
\texttt{\quad "edit\_type": "removal",} \\
\texttt{\quad "instruction": "Draw the result that half of the apples on the right-side of the desk are eaten and half of the apples left on the desk.",} \\
\texttt{\quad "edited\_effect": "There are only 2 apples left on the desk."} \\
\texttt{\}} \\

\textbf{Example 4 (Size + Change Color)} \\
* \textbf{Context:} User asks ``Which of the three suitcases is the largest?'' and GPT answers ``The red one on the left.'' \\
* \textbf{Output:} \\
\texttt{\{} \\
\texttt{\quad "reasoning\_type": "size",} \\
\texttt{\quad "edit\_target": "the largest suitcase",} \\
\texttt{\quad "edit\_type": "change color",} \\
\texttt{\quad "instruction": "Identify the largest piece of luggage in the group and change its color to bright blue.",} \\
\texttt{\quad "edited\_effect": "The largest suitcase, originally red, is now bright blue."} \\
\texttt{\}} \\

\tcbbreak
\textbf{Example 5 (Distance + Removal)} \\
* \textbf{Context:} User asks ``Which person is standing furthest away from the camera?'' and GPT answers ``The man in the yellow jacket.'' \\
* \textbf{Output:} \\
\texttt{\{} \\
\texttt{\quad "reasoning\_type": "distance",} \\
\texttt{\quad "edit\_target": "the man furthest away",} \\
\texttt{\quad "edit\_type": "removal",} \\

\texttt{\quad "instruction": "Identify the person standing at the greatest distance from the viewer and remove them from the scene.",} \\
\texttt{\quad "edited\_effect": "The man in the yellow jacket in the far distance is removed, leaving the background scenery empty."} \\
\texttt{\}} \\
\textbf{Example 6 (Depth + Replace with similar-sized object)} \\
* \textbf{Context:} User asks ``What is in the foreground, blocking the view of the house?'' and GPT answers ``A large oak tree.'' \\
* \textbf{Output:} \\
\texttt{\{} \\
\texttt{\quad "reasoning\_type": "depth",} \\
\texttt{\quad "edit\_target": "the oak tree in the foreground",} \\
\texttt{\quad "edit\_type": "replace with similar-sized object",} \\
\texttt{\quad "instruction": "Identify the large plant in the immediate foreground that obscures the building, and replace it with a tall, slender pine tree of similar height.",} \\
\texttt{\quad "edited\_effect": "The large oak tree in the foreground is replaced by a tall pine tree, maintaining the depth of the scene."} \\
\texttt{\}} \\

\textbf{Example 7 (Temporal + Future Prediction)} \\
* \textbf{Context:} User asks ``What will happen to the ice cream cone sitting in the hot sun?'' and GPT answers ``It will melt.'' \\
* \textbf{Output:} \\
\texttt{\{} \\
\texttt{\quad "reasoning\_type": "temporal",} \\
\texttt{\quad "edit\_target": "the ice cream cone",} \\
\texttt{\quad "edit\_type": "future prediction",} \\
\texttt{\quad "instruction": "Based on the weather conditions, draw the state of the dessert sitting on the table 10 minutes later.",} \\
\texttt{\quad "edited\_effect": "The ice cream cone is completely melted into a puddle on the table."} \\
\texttt{\}} \\
\textbf{Output Format (JSON Only):} \\
\texttt{\{} \\
\texttt{\quad "reasoning\_type": "...",} \\
\texttt{\quad "edit\_target": "...",} \\
\texttt{\quad "edit\_type": "...",} \\
\texttt{\quad "instruction": "...",} \\
\texttt{\quad "edited\_effect": "..."} \\
\texttt{\}}

\end{promptbox}